\documentclass[12pt,a4paper]{article}
\usepackage[a4paper,left=2.5cm,right=2.5cm,top=2.5cm,bottom=2.5cm]{geometry}

\usepackage{algorithm}
\usepackage[noend]{algpseudocode}
\usepackage{amsmath}
\usepackage{float}
\usepackage[title]{appendix}
\usepackage{adjustbox,lipsum}
\usepackage{url}
\usepackage{threeparttable}
\usepackage{subfig}
\usepackage{graphicx}

\DeclareMathOperator*{\argmax}{argmax}
\algdef{SE}[SUBALG]{Indent}{EndIndent}{}{\algorithmicend\ }%
\algtext*{Indent}
\algtext*{EndIndent}

\begin{document}
\date{}
\title{Adapting multi-armed bandits policies to contextual bandits scenarios}
\author{David Cortes}
\maketitle

\begin{abstract}
This work explores adaptations of successful multi-armed bandits policies to the online contextual bandits scenario with binary rewards using binary classification algorithms such as logistic regression as black-box oracles. Some of these adaptations are achieved through bootstrapping or approximate bootstrapping, while others rely on other forms of randomness, resulting in more scalable approaches than previous works, and the ability to work with any type of classification algorithm. In particular, the \textsc{Adaptive-Greedy} algorithm (\cite{mortalmab}) shows a lot of promise, in many cases achieving better performance than upper confidence bound and Thompson sampling strategies, at the expense of more hyperparameters to tune.
\end{abstract}

\section{Introduction}

Contextual bandits, also known as associative reinforcement learning (\cite{linucb0}) or multi-armed bandits with covariates (\cite{mbwcov}), is a problem characterized by the following iterative process: there is a number of choices (known as "arms") from which an agent can choose, which contain stochastic rewards. At the beginning of each round, the world generates a set of covariates of fixed dimensionality (so-called "context"), and rewards for each arm which are related to the covariates. The agent chooses an arm for that round, and the world reveals the reward for that arm, but not for the others. The goal is for the agent to maximize its obtained rewards in the long term, using his previous actions' history.

This is related to the simpler multi-armed bandits problem (MAB, \cite{mablong}), which has no covariates, but faces the same dilemma between exploration of unkown alternatives or exploitation of known good arms, for which many approaches have been proposed such as upper confidence bounds (\cite{mablong}, \cite{linucb0}), Thompson sampling (\cite{ts}) or other heuristics (\cite{poker}).

This work proposes adaptations of some successful strategies and baselines that have been proposed for the MAB setting and variations thereof to the contextual bandits setting by using supervised learning algorithms as black-box oracles, as well as other considerations such as exploration in the early phases in the absence of non-zero rewards, benchmarking them in an empirical evaluation using multilabel classification datasets with different characteristics from which the studied problem is simulated as in \cite{doublyrobust}.

\section{Problem defintion}

More formally, this work is concerned with a scenario as follows: there is a fixed number of choices or arms $k$, from which an agent must choose one as his action $a_t$ in each round $t$. At the beginning of each round, the world generates a set of covariates $\mathbf{x}_m^t$ (a.k.a. the "context") of fixed dimensionality $m$, and stochastic binary rewards for each arm $\mathbf{r}_k^t \in \{ 0, 1 \}$ through a function of the covariates $r_k^t \sim \text{Bernoulli}(f_k(\mathbf{x}^t))$ which is different for each arm but is the same throughout all rounds. The world reveals $\mathbf{x}_m^t$ to the agent, which must then choose an arm based on his previous knowledge, and the reward for the arm that was chosen is revealed, while the rewards for the other arms remain unknown. The agent shall use his history of previously seen covariates $\mathcal{X}$, chosen actions $\mathcal{A}$ and observed rewards $\mathcal{R}$ during each round in order to make a choice. Note that this work is only concerned with Bernoulli-distributed rewards, which is a setting encountered in domains such as recommender systems or online advertising in which a user either clicks or doesn't click what she is presented with.

The objective is to maximize the rewards obtained in the long term, without a defined time limit. Compared to MAB which evaluates arm-selection policies based on upper bounds on regret (\cite{mablong}), defined there as the difference between the reward from the arm selected at each round and the highest expected reward of any arm, this a less clear objective in online contextual bandits. Alternative definitions of regret based on the true reward-generating function have been proposed (\cite{linucb0}), and regret bounds have been studied for the case of linear functions. Other works have also tried to define regret differently, such as \cite{exp4}, but their definitions are not applicable in this scenario. While some methods that enjoy theoretical guarantees on their regret have been proposed before for contextual bandits, either using another algorithm as oracle or not, such methods tend to be computationally intractable such as \cite{monster}, \cite{onlinecover}, or \cite{lipshitz}, and this work tries to explore more practical and scalable approaches at the expense of theoretical guarantees.

Alternative objectives such as sums of weighted rewards discounting later gains have also been proposed in the MAB setting, but this does not necessarily result in a good definition of "long term", and is subject to variations in the discount rate used.

As such, this work is concerned with cumulative reward throughout the rounds instead of accumulated regret, with time periods running up to the available number of rows in the given datasets. While this has some chance of not being able to reflect asymptotic behaviors in an infinite-time scenario with all-fixed arms, it provides some insight on what happens during typical timelines of interest.

\section{Related work}

The simpler multi-armed bandits scenario has been extensively studied, and many good solutions have been proposed which enjoy theoretical limits on their regrets (\cite{mablong}), as well as demonstrated performance in empirical tests (\cite{poker}). Among the solutions with best theoretical guarantees and empirical performance are \emph{upper confidence bounds}, also known as "optimism in the face of uncertainty", which try to establish an upper bound on the expected reward (such bound gets closer to the observed mean as more observations are accumulated, thereby balancing exploration and exploitation), and \emph{Thompson sampling} which takes a Bayesian perspective aiming to choose an arm according to its probability of being the best arm. Typical comparison baselines are \emph{Epsilon-Greedy} algorithms and variations thereof, whose idea is to select the empirical best action with some probability or a random action otherwise. In the time-limited setting, other logical strategies have also been evaluated, such as choosing random actions at the beginning but then shifting to always playing the empirically best action after an optimal turning point (known as \emph{Explore-Then-Exploit}).

Variations of the multi-armed bandit setting have also seen other interesting proposals, such as the \textsc{Adaptive-Greedy} algorithm proposed for the mortal (expiring) arms case in \cite{mortalmab}, which selects the empirically best arm if its estimated expected reward is above a certain threshold, or a random arm otherwise.

The contextual bandits setting has been studied as different variations of the problem formulation, some differing a lot from the one presented here such as the bandits with "expert advise" in \cite{exp3} and \cite{exp4}, and some presenting a similar scenario in which the rewards are assumed to be continuous (usually in the range $ [ 0, 1 ] $) and the reward-generating functions linear (\cite{linucb1}, \cite{linucb2}). Particularly, \textsc{LinUCB} (\cite{linucb1}), which as its name suggests uses a linear function estimator with an upper bound on the expected rewards (one estimator per arm, all independent of each other), has proved to be a popular approach and many works build upon it in variations of its proposed scenario, such as when adding similarity information (\cite{gang}, \cite{colin}).

Approaches taking a supervised learning algorithm as an oracle for a similar setting as presented here but with continuous rewards have been studied before (\cite{bsts}, \cite{langford}), in which these oracles are fit to the covariates and rewards from each arm separately, and the same strategies from multi-armed bandits have also resulted in good strategies in this setting. Other related problems such as building an optimal oracle or policy with data collected from a past policy have also been studied (\cite{offsettree}, \cite{dr2}, \cite{onlinecover}), but this work only focuses on online policies that start from scratch and continue ad-infinitum.

Another related problem that has been deeply studied is \emph{active learning} (\cite{activelearning}), which assumes a scenario in which a supervised learning algorithm must make predictions about many observations, but revealing the true label of an observation in order to incorporate it in its training repertoire is costly, thus it must actively select which observations to label in order to improve predictions. In the classification scenario with differentiable models, a simple yet powerful technique that has been tried is to look at the gradient that an observation would have on the loss or likelihood function if its true label were known (there are only two possible outcomes in this case, and some methods provide a probability of each being the correct one), following the idea that larger gradients e.g. as measured by some vector norm, will lead to faster learning.

\section{Obtaining upper confidence bounds}

The aim of upper confidence bounds for an oracle (fit to rewards and covariates coming from a single arm) is to be able to establish a bound for the expected reward of an arm given the covariates or features, below which lies the true expected reward with a high probability. The tighter the bound at the same probability, the better. As the number of observations grows, this bound should get closer to the point estimate of the expected reward generated by the oracle.

Previous works have tried to use the upper confidence bound strategy in contextual bandits by upper-bounding the standard error of predictions under assumptions on the reward-generating functions such as \cite{linucb0} and \cite{linucb1}. However, methods typically used in the statistics and econometrics literature for the same purpose have been overlooked, such as Bayesian sampling (\cite{nuts}, \cite{varinf}), estimations of predictor's covariance under normality assumptions (\cite{vcov}), and bootstrapping (\cite{bootstrap}), which can result in tighter upper confidence bounds and more scalable approaches than \cite{linucb0} and \cite{linucb1}.

The Bayesian approach would restrict the oracles to certain classes of models and might not result in a very scalable strategy (albeit stochastic variational inference might in some cases result in an online and fast-enough procedure), while the covariance estimations restrict the class of oracles to generalized linear models, but the bootstrapping approach results in a very scalable strategy that can work with any class of supervised learning algorithms (including collaborative filtering ones) and which makes almost no assumptions on the reward-generating functions.

To recall, bootstrapping consists in taking resamples of the data of the same size as the original by picking observations at random from the available pool but with replacement. These resamples are drawn from the same distribution as the originals, thus can be used to construct confidence intervals of parameters or other statistics of interest, such as the estimation of the expected reward for an arm given its covariates, by simply taking the statistic of interest estimated under each resample and calculating quantiles on it.

Obtaining upper confidence bounds this way is straightforward, but it has one inconvenience: it requires access to the whole dataset. In some cases, it might be desirable to have oracles that use online learning methods (i.e. stochastic optimization), which are fit to online data streams or small batches of observations incrementally instead of being refit to the whole data every time. In theory, as the sample size grows to infinity, the number of times that an observation appears in a resample should be a random number distributed $\sim \text{Poisson}(1)$ (\cite{onlinebs}), and one possibility is to take each observation as it arrives a random number of times $n \sim \text{Poisson}(1)$. Alternative approaches have also been proposed, such as \cite{bts}, which use a different distribution for the number of times that an observation appears. This work found a more practical but less theoretically correct method for dealing with this problem: assigning sample weights at random $\sim \text{Gamma}(1,1)$, which are passed to the classification oracle provided that it supports them. This produces a more stable effect at smaller sample sizes, as it avoids the problem of ending up with observations that have only one label. For more information on it see the appendix.

For some classes of algorithms such as decision trees, it should be possible to calculate an upper bound also by looking at the data on the terminal nodes, and methods such as random forests that implicitly perform bootstrapping should be able to produce an upper confidence bound without additional bootstrapping. Such methods however were not explored in this work.

\section{Cold-start problem}

One issue that makes the contextual bandits scenario harder is that most supervised learning algorithms used as oracles cannot be fit to data that has only one value or only one label (e.g. only observations which had no reward), and typical domains of interest involve a scenario in which the non-zero reward rate for any arm is rather small regardless of the covariates (e.g. clicks). In the MAB setting, this is usually solved by incorporating a prior or some smoothing criterion, and it's possible to think of a similar fix for the scenario proposed in this work if the classifier is able to output probabilities: if all the observations for a given arm have the same label/outcome, always predict that label for that arm, but add a smoothing criterion regardless:
$$ \hat{r}_{smooth} = \frac{n \times \hat{r} + a}{n + b} $$
where $\hat{r} \in [0, 1 ]$ is the expected reward estimated by the oracle, $n$ is the number of observations for that arm, and $a, b$ with $a < b$ are smoothing constants. One might also think of incorporating artificial observations with an unseen label, but this can end up doing more harm than good. A recalibration can also be applied to the outputs of classifiers that don't produce probabilities in order to make this heuristic work (\cite{kuhn}).

However, as the arm sizes grow larger and the problem starts resembling more the \emph{many-armed bandits} scenario (\cite{manyarmed1}, \cite{manyarmed2}), in which there might be more arms than rounds or time steps, it's easy to see that this smoothing criterion will lead to pretty much sampling each arm once or twice, which in turn will lead to low rewards until all arms have been sampled a certain number of times. Taking this into consideration, another logical solution would be to start with a Bayesian multi-armed bandit policy for each arm that uses a Beta prior and ignores the covariates, then switch to a contextual bandit policy once a minimum number of observations from each label has been obtained for that arm (as there is randomness involved, it is highly unlikely that it will start by sampling each arm as the smoothing would do in a mostly-zero reward scenario).

\begin{algorithm}[H]
\caption{MAB-first}\label{MAB-first}
\hspace*{\algorithmicindent} \textbf{Inputs} const. $a, b$, threshold $m$, contextual bandit policy $\pi_k$, covariates $\mathbf{x}$ \\
\hspace*{\algorithmicindent} \textbf{Output} score for arm $\hat{r}_k$
\begin{algorithmic}[1]
\If {$| \{ r \in \mathcal{R}_k \: | \: r = 0 \} | <m$ or $| \{r \in \mathcal{R}_k \: | \: r = 1 \} | <m $} 
	\State Sample $\hat{r}_k \sim \text{Beta}(a + | \{r \in \mathcal{R}_k | r = 1 \} |,\: b + |\{ r \in \mathcal{R}_k | r = 0 \} |)$
\Else \State Set $\hat{r}_k = \pi_k(x)$
\EndIf
\Return $\hat{r}_k$
\end{algorithmic}
\end{algorithm}

The proposed algorithms were evaluated by complementing them with both the smoothing technique and the \textsc{MAB-first} technique, which usually proved to be a better choice. For more information on this comparison, see the appendix. Some generic suggestion for these constants are as follows: $a = 1, b = 1, m = 2$; $a = 3, b = 7, m = 3$; $a = \frac{5}{k}, b = 4, m = 2$.

Another logical solution for the problem of having too many arms is of course to limit oneself to a randomly chosen subset of them, perhaps adding more with time, but it's hard to determine what would be the optimal number given some timeframe, especially in situations in which arms expire and/or new arms are available later. See the appendix for such comparison.

An idea to incorporate more observations into arms is to add each observation which results in a reward for one arm as a non-reward observation for all other arms. In scenarios in which only one arm per round tends to have a reward, or mostly one arm per round only, this can provide a small lift, but in scenarios in which this is not the case, it can do more harm in the long term.

\section{Establishing baselines}

Before determining the quality of a policy or strategy for contextual bandits, it's a good idea to establish simple comparison points that any good policy or strategy should be able to beat in order to ensure that it is indeed a good policy.

MAB has also seen many proposals in this regard, with the most simple one being Epsilon-Greedy algorithms, which consist in playing the empirical best arm with some high probability or a random arm otherwise. Variations of it have also been proposed, such as decreasing the probability of choosing a random arm with each successive round, or dropping the probability of picking a random arm to zero after some turning point. This algorithm lends itself to an easy adaptation to the problem studied here:

\begin{algorithm}[H]
\caption{Epsilon-Greedy} \label{Epsilon-Greedy}
\hspace*{\algorithmicindent} \textbf{Inputs} probability $p \in (0, 1]$, decay rate $d \in (0, 1] $, oracles $\hat{f}_{1:k}$ \
\begin{algorithmic}[1]
\For {each succesive round $t$ with context $\mathbf{x}^t$}

	\State With probability $(1 - p)$:
    \Indent
         \State Select action $a = \argmax_k \hat{f}_{k}(\mathbf{x}^t)$
    \EndIndent
    \State Otherwise:
    \Indent
         \State Select action $a$ uniformly at random from $1$ to $k$
    \EndIndent

	\State Update $p := p \times d$
	\State Obtain reward $r_a^t$, Add observation $\{ \mathbf{x}^t, r_a^t \}$ to the history for arm $a$
	\State Update oracle $\hat{f}_a$ with its new history
	
\EndFor
\end{algorithmic}
\end{algorithm}

The oracles $\hat{f}_{1:k}$ would consist of separate and independent binary classifiers (such as XGBoost or logistic regression), each fit only to the observations and rewards from the rounds in which its respective arm was chosen, wrapped inside the \textsc{MAB-first} or the smoothing criterion as described in the previous section ($\pi_k(\mathbf{x}) = \hat{f}_k^{orig}(\mathbf{x})$), and ties broken arbitrarily. While ideally the oracles should be updated after every iteration, and stochastic optimization techniques allow doing this for many classification algorithms, they might also be updated only after a certain amount of rounds, or every time each one's history accumulates a certain number of new cases, at the expense of some decrease in the predictive power of the oracles - this becomes less of an issue as the histories' lenght increases.

Following the idea of alternating between exploring new alternatives under new contexts and exploiting the accrued knowledge from previous exploration, another logical option is to alternate between periods of choosing actions at random, and periods of playing the actions with the highest estimated expected reward (a more extreme case of the \textsc{Epoch-Greedy} policy proposed in \cite{epochgreedy}). In the most extreme case, if there is a timeline defined, this strategy would consist of playing an arm at random up to an optimal turning point, after which it will only play the arm with highest expected reward.

\begin{algorithm}[H]
\caption{Explore-Then-Exploit} \label{Explore-Then-Exploit}
\hspace*{\algorithmicindent} \textbf{Inputs} breakpoint $t_b$, oracles $\hat{f}_{1:k}$ \
\begin{algorithmic}[1]
\For {each succesive round $t$ with context $\mathbf{x}^t$}

	\If {$t < t_b$}
		\State Select action $a$ uniformly at random from $1$ to $k$
	\Else
		\State Select action $a = \argmax_k \hat{f}_k(\mathbf{x}^t)$
	\EndIf

	\State Obtain reward $r_a^t$, Add observation $\{ \mathbf{x}^t, r_a^t \}$ to the history for arm $a$
	\State Update oracle $\hat{f}_a$ with its new history
	
\EndFor
\end{algorithmic}
\end{algorithm}

In practice, we don't have a pre-defined time limit, but since the sample size is known beforehand when using existing datasets, it can be added as another baseline.

\sloppy Another logical idea that has been used in other problem domains when making a decision from uncertain estimations is to choose not by a simple $\argmax$, but with a probability proportional to the estimates, e.g. $ a \sim \text{Mult}(\text{softmax}(\hat{f}_1(\mathbf{x}), ..., \hat{f}_k(\mathbf{x}))) $, where $\text{softmax}(\mathbf{x}_1, ..., \mathbf{x}_n) = \exp(\mathbf{x}_1, ..., \mathbf{x}_n) / \sum_{n} \exp(\mathbf{x}_n) $. As the estimates here are probabilities bounded between zero and one, it might make more sense to apply an inverse sigmoid function $\text{sigmoid}^{-1}(\mathbf{x}) = \log(\frac{\mathbf{x}}{1 - \mathbf{x}})$ on these probabilities before applying the softmax function. In order to make such policy converge towards an optimal strategy in the long-term, a typical trick is to inflate the estimates before applying the softmax function by a multiplier that gets larger with the number of rounds, so that the policy would tend to $\argmax$ with later iterations.

\begin{algorithm}[H]
\caption{SoftmaxExplorer} \label{SoftmaxExplorer}
\hspace*{\algorithmicindent} \textbf{Inputs} oracles $\hat{f}_{1:k}$, multiplier $m$, inflation rate $i$ \
\begin{algorithmic}[1]
\For {each succesive round $t$ with context $\mathbf{x}^t$}

	\State Sample action $a \sim \text{Mult}(\text{softmax}(m \times \text{sigmoid}^{-1}(\hat{f}_1(\mathbf{x}^t), ..., \hat{f}_k(\mathbf{x}^t))))$
	\State Update $m := m \times i$
	\State Obtain reward $r_a^t$, Add observation $\{ \mathbf{x}^t, r_a^t \}$ to the history for arm $a$
	\State Update oracle $\hat{f}_a$ with its new history
	
\EndFor
\end{algorithmic}
\end{algorithm}

Finally, another good baseline is to always select the arm with the highest average reward ignoring the context. A good MAB policy should quickly converge towards always choosing such arm.

\section{Algorithms}

Following the previous sections, a natural adaptation of the upper confidence bound strategy is as follows:

\begin{algorithm}[H]
\caption{BootstrappedUCB} \label{BootstrappedUCB}
\hspace*{\algorithmicindent} \textbf{Inputs} number of resamples $m$, percentile $p$, oracles $\hat{f}_{1:k,1:m}$ \
\begin{algorithmic}[1]
\For {each succesive round $t$ with context $\mathbf{x}^t$}
	\For {arm $q$ in $1$ to $k$}
		\State Set $\hat{r}_{q}^{ucb} = \text{Percentile}_p \{ \hat{f}_{q, 1}(\mathbf{x}^t), ..., \hat{f}_{q, m}(\mathbf{x}^t) \} $
	\EndFor
	
	\State Select action $a =  \argmax_{q} \hat{r}_{q}^{ucb}$
	
	\State Obtain reward $r_a^t$, Add observation $\{ \mathbf{x}^t, r_a^t \}$ to the history for arm $a$
	
	\For {resample $s$ in $1$ to $m$}
		\State Take bootstrapped resample $\mathbf{X}_s, \mathbf{r}_s$ from $\mathcal{X}_a, \mathcal{R}_a$
		\State Refit $\hat{f}_{a, s}$ to this resample
	\EndFor
	
\EndFor
\end{algorithmic}
\end{algorithm}

And its online variant:

\begin{algorithm}[H]
\caption{OnlineBoostrappedUCB} \label{OnlineBoostrappedUCB}
\hspace*{\algorithmicindent} \textbf{Inputs} number of resamples $m$, percentile $p$, oracles $\hat{f}_{1:k,1:m}$ \
\begin{algorithmic}[1]
\For {each succesive round $t$ with context $\mathbf{x}^t$}
	\For {arm $q$ in $1$ to $k$}
		\State Set $\hat{r}_{q}^{ucb} = \text{Percentile}_p \{ \hat{f}_{q, 1}(\mathbf{x}^t), ..., \hat{f}_{q, m}(\mathbf{x}^t) \} $
	\EndFor
	
	\State Select action $a =  \argmax_{q} \hat{r}_{q}^{ucb}$
	
	\State Obtain reward $r_a^t$, Add observation $\{ \mathbf{x}^t, r_a^t \}$ to the history for arm $a$
	
	\For {resample $s$ in $1$ to $m$}
		\State Sample observation weight $w \sim \text{Gamma}(1, 1)$
		\State Update $\hat{f}_{a, s}$ with the new observation $\{ \mathbf{x}^t, r_a^t \}$ with weight $w$
	\EndFor
	
\EndFor
\end{algorithmic}
\end{algorithm}

Just like before, the oracles (binary classifiers) might not be updated after every new observation is incorporated, but only after a reasonable number of them has been incorporated, or after a fixed number of rounds.

Thompson sampling results in an even more straight forward adaptation:

\begin{algorithm}[H]
\caption{BootstrappedTS} \label{BootstrappedTS}
\hspace*{\algorithmicindent} \textbf{Inputs} number of resamples $m$, oracles $\hat{f}_{1:k,1:m}$ \
\begin{algorithmic}[1]
\For {each succesive round $t$ with context $\mathbf{x}^t$}
	\For {arm $q$ in $1$ to $k$}
		\State Select resample $s$ uniformly at random from $1$ to $m$
		\State Set $\hat{r}_{q}^{ts} = \hat{f}_{q,s}(\mathbf{x}^t)$
	\EndFor
	
	\State Select action $a = \argmax_q \hat{r}_{q}^{ts}$
	
	\State Obtain reward $r_a^t$, Add observation $\{ \mathbf{x}^t, r_a^t \}$ to the history for arm $a$
	
	\For {resample $s$ in $1$ to $m$}
		\State Take bootstrapped resample $\mathbf{X}_s, \mathbf{r}_s$ from $\mathcal{X}_a, \mathcal{R}_a$
		\State Refit $\hat{f}_{a, s}$ to this resample
	\EndFor
	
\EndFor
\end{algorithmic}
\end{algorithm}

And similarly, its online variant:

\begin{algorithm}[H]
\caption{OnlineBootstrappedTS} \label{OnlineBootstrappedTS}
\hspace*{\algorithmicindent} \textbf{Inputs} number of resamples $m$, oracles $\hat{f}_{1:k,1:m}$ \
\begin{algorithmic}[1]
\For {each succesive round $t$ with context $\mathbf{x}^t$}
	\For {arm $q$ in $1$ to $k$}
		\State Select resample $s$ uniformly at random from $1$ to $m$
		\State Set $\hat{r}_{q}^{ts} = \hat{f}_{q,s}(\mathbf{x}^t)$
	\EndFor
	
	\State Select action $a = \argmax_q \hat{r}_{q}^{ts}$
	
	\State Obtain reward $r_a^t$, Add observation $\{ \mathbf{x}^t, r_a^t \}$ to the history for arm $a$
	
	\For {resample $s$ in $1$ to $m$}
		\State Sample observation weight $w \sim \text{Gamma}(1, 1)$
		\State Update $\hat{f}_{a, s}$ with the new observation $\{ \mathbf{x}^t, r_a^t \}$ with weight $w$
	\EndFor
	
\EndFor
\end{algorithmic}
\end{algorithm}

Outside of algorithms relying on bootstrapping, some algorithms that use a random selection criterion  also result in easy adaptations with reasonably good performance without requiring multiple oracles per arm, such as the \textsc{Adaptive-Greedy} (\cite{mortalmab}) algorithm:

\begin{algorithm}[H]
\caption{ContextualAdaptiveGreedy} \label{ContextualAdaptiveGreedy}
\hspace*{\algorithmicindent} \textbf{Inputs} threshold $z \in (0,1)$, decay rate $d \in (0, 1]$, oracles $\hat{f}_{1:k}$ \
\begin{algorithmic}[1]
\For {each succesive round $t$ with context $\mathbf{x}^t$}
	
	\If {$\max_{k} \hat{f}_k(\mathbf{x}^t) > z$}
		\State Select action $a = \argmax_k \hat{f}_{k}(\mathbf{x}^t)$
	\Else
		\State Select action $a$ uniformly at random from $1$ to $k$
	\EndIf
	
	\State Update $ z := z \times d$
	\State Obtain reward $r_a^t$, Add observation $\{ \mathbf{x}^t, r_a^t \}$ to the history for arm $a$
	\State Update oracle $\hat{f}_a$ with its new history
	
\EndFor
\end{algorithmic}
\end{algorithm}

The choice of threshold $z$ is problematic though, and it might be a better idea to base it instead on the estimations produced by the oracles - for example by keeping a moving average window of the last $m$ highest estimated rewards of the best arm:

\begin{algorithm}[H]
\caption{ContextualAdaptiveGreedy2} \label{ContextualAdaptiveGreedy2}
\hspace*{\algorithmicindent} \textbf{Inputs} window size $m$, initial threshold $z_0$, decay $d \in (0,1] $, oracles $\hat{f}_{1:k}$ \
\begin{algorithmic}[1]
\For {each succesive round $t$ with context $\mathbf{x}^t$}
	
	\If {$t = 1$}
		\State Set $z = z_0$
	\EndIf	
	
	\State Set $\hat{r}_t = \max_{k} \hat{f}_k(\mathbf{x}^t)$
	
	\If {$\hat{r}_t > z$}
		\State Select action $a = \argmax_k \hat{f}_{k}(\mathbf{x}^t)$
	\Else
		\State Select action $a$ uniformly at random from $1$ to $k$
	\EndIf
	
	\If {$t \ge m$}
		\State Update $ z := \text{Percentile}_p \{ \hat{r}_{t}, \hat{r}_{t-1}, ..., \hat{r}_{t-m+1} \} $
		\State Update $ p := p \times d$
	\EndIf
	\State Obtain reward $r_a^t$, Add observation $\{ \mathbf{x}^t, r_a^t \}$ to the history for arm $a$
	\State Update oracle $\hat{f}_a$ with its new history
	
\EndFor
\end{algorithmic}
\end{algorithm}

This moving window in turn might also be replaced with a non-moving window, i.e. compute the average for the first $m$ observations, but don't update it until $m$ more rounds, then at time $2m$ update only with the observations that were between $m$ and $2m$.

Instead of relying on choosing arms at random for exploration, active learning heuristics might be chosen for faster learning instead. Strategies such as \textsc{Epsilon-Greedy} are easy to convert into active learning - for example, assuming a differentiable and smooth model such as logistic regression or artificial neural networks (depending on the particular activation functions):

\begin{algorithm}[H]
\caption{ActiveExplorer} \label{ActiveExplorer}
\hspace*{\algorithmicindent} \textbf{Inputs} probability $p$, oracles $\hat{f}_{1:k}$, gradient functions for oracles $g_{1:k}(\mathbf{x}, r)$ \
\begin{algorithmic}[1]
\For {each succesive round $t$ with context $\mathbf{x}^t$}

	\State With probability $p$:
    \Indent
         \State Select action $a = \argmax_k \hat{f}_{k}(\mathbf{x}^t)$
    \EndIndent
    \State Otherwise:
    \Indent
         \For {arm $q$ in $1$ to $k$}
         	\State Set $z_q = (1 - \hat{f}_q(\mathbf{x}^t)) \lVert g_q (\mathbf{x}^t, 0) \lVert \: + \: \hat{f}_q(\mathbf{x}^t) \lVert g_q (\mathbf{x}^t, 1) \lVert$
         \EndFor
         \State Select action $a = \argmax_k z_k$
    \EndIndent

	\State Obtain reward $r_a^t$, Add observation $\{ \mathbf{x}^t, r_a^t \}$ to the history for arm $a$
	\State Update oracle $\hat{f}_a$ with its new history, along with $\hat{g}_a$
	
\EndFor
\end{algorithmic}
\end{algorithm}

Intuitively, it might also be a good idea to take the arm with the smallest or largest gradient for either label instead of a weighted average according to the estimated probabilities, as each alternative (max, min, weighted) seeks something that adds value, but in practice a weighted average tends to give slightly better results. For a comparison see the appendix.

The previously defined \textsc{ContextualAdaptiveGreedy2} for example can also be enriched with this simple heuristic:

\begin{algorithm}[H]
\caption{ActiveAdaptiveGreedy} \label{ActiveAdaptiveGreedy}
\hspace*{\algorithmicindent} \textbf{Inputs} window size $m$, initial threshold $z_0$, decay rate $d \in (0,1]$, oracles $\hat{f}_{1:k}$, gradient functions for oracles $g_{1:k}(\mathbf{x}, r)$ \
\begin{algorithmic}[1]
\For {each succesive round $t$ with context $\mathbf{x}^t$}
	
	\If {$t = 1$}
		\State Set $z = z_0$
	\EndIf	
	
	\State Set $\hat{r}_t = \max_{k} \hat{f}_k(\mathbf{x}^t)$
	
	\If {$\hat{r}_t > z$}
		\State Select action $a = \argmax_k \hat{f}_{k}(\mathbf{x}^t)$
	\Else
		\For {arm $q$ in $1$ to $k$}
			\State Set $z_q = (1 - \hat{f}_q(\mathbf{x}^t)) \lVert g_q (\mathbf{x}^t, 0) \lVert \: + \: \hat{f}_q(\mathbf{x}^t) \lVert g_q (\mathbf{x}^t, 1) \lVert$
		\EndFor
		\State Select action $a = \argmax_k z_k$
	\EndIf
	
	\If {$t \ge m$}
		\State Update $ z := \text{Percentile}_p \{ \hat{r}_{t}, \hat{r}_{t-1}, ..., \hat{r}_{t-m+1} \} $
	\EndIf
	\State Obtain reward $r_a^t$, Add observation $\{ \mathbf{x}^t, r_a^t \}$ to the history for arm $a$
	\State Update oracle $\hat{f}_a$ with its new history, along with $\hat{g}_a$
	
\EndFor
\end{algorithmic}
\end{algorithm}

\section{Empirical evaluation}

The algorithms above (implementations are open-source and freely available\footnote{\url{https://github.com/david-cortes/contextualbandits}}) were benchmarked and compared to the simpler baselines by simulating contextual bandits scenarios using multi-label classification datasets, where the arms become the classes and the rewards are whether the chosen label for a given observation was correct or not. This was done by feeding them observations in rounds, letting each algorithm make its choice but presenting the same context to all, and revealing to each one whether the label that it chose in that round was correct or not.

The datasets used are the BibTeX tags (\cite{bibtex}), Del.icio.us tags (\cite{delicious}), Mediamill (\cite{mediamill}), and EURLex\footnote{Due to speed reasons, this dataset was only used for the simulations in the appendix B} (\cite{eurlex}) (these were taken from the Extreme Classification Repository\footnote{\url{http://manikvarma.org/downloads/XC/XMLRepository.html}}), representing a variety of problem domains and datasets with different properties, such as having a dominant arm to which most observations belong (Mediamill), having a large number of labels in relation to the number of available rounds with some never offering a reward (EURLex), or a more balanced scenario with no dominant label (BibTeX). The same Mediamill dataset was also used without the 5 most common labels, which results in a very different scenario. The dataset sizes, average number of labels per row, average number of rows per label, and percent of observations having the most common label are as follows:

\begin{adjustbox}{max width=\textwidth}{\centering
\begin{tabular}{|l|c|c|c|c|c|c|}
 \hline
  & \textbf{Obs.} & \textbf{Feats.} & \textbf{Labels} & \textbf{Lab./obs.} & \textbf{Obs/lab.} & \textbf{Most common} \\
 \hline
\textbf{BibTeX} & 7,395 & 1,836 & 159 & 2.4 & 111.71 & 14.09\% \\ \hline
\textbf{Del.icio.us} & 16,105 & 500 & 983 & 19.02 & 311.61 & 40.33\% \\ \hline
\textbf{MediaMill} & 43,907 & 120 & 101 & 4.38 & 1902.16 & 77.14\% \\ \hline
\textbf{MediaMill Reduced} & 43,907 & 120 & 96 & 2.07 & 945.1 & 17.56\% \\ \hline
\textbf{EURLex} & 19,348 & 5,000 & 3,956 & 5.31 & 25.79 & 6.48\% \\ \hline
 \hline
\end{tabular}}\end{adjustbox}\\

The classification oracles were refit every 50 rounds, and the experiments were run until iterating throughout all observations in a dataset, after which it was done again with the data shuffled differently, and the results averaged over 10 runs. Both full-refit and mini-batch-update versions were evaluated. The classifier algorithm used was logistic regression, with the same regularization parameter for every arm.

Contextual bandits policies were evaluated by their plots of cumulative mean reward over time (that is, the average reward per round obtained up to a given round), with time being the number of rounds or observations, and the reward being whether they choose a correct label (arm) for an observation (context).

No feature engineering or dimensionality reduction was performed, as the point was to compare metaheuristics.

Unfortunately, algorithms such as \textsc{LinUCB} don't scale to these problem sizes, so it was not possible to compare against them. Additionally, they are intended for the case of continuous rather than binary rewards, and as such their performance might not be as good for this binary rewards scenario.

The \textsc{MAB-first} technique was used in all cases except for \textsc{Explore-Then-Exploit}, but its hyperparameters were not tuned, nor were the hyperparameters of the contextual bandit policies. The results in the appendix suggest that good tuning of the \textsc{MAB-first} hyperparameters can have a large impact. The hyperparameters were $a=3, b=7, m=2$ for the BibTeX and Del.icio.us datasets, and $a=1, b=10, m=2$ for the Mediamill datasets.

The other policies' hyperparameters were set as follows: 10 resamples for bootstrapped methods, 80\% confidence interval for UCB, 20\% explore probability for \textsc{Epsilon-Greedy} and 15\% for \textsc{ActiveExplorer}, decay rate $0.9999$ for \textsc{Epsion-Greedy}, $0.9997$ for \textsc{ContextualAdaptiveGreedy}, \textsc{ContextualAdaptiveGreedy2} and \textsc{AdaptiveActiveGreedy}, multiplier of $2.0$ for \textsc{SoftmaxExplorer} with inflation rate $1.001$, percentile 30 for \textsc{ContextualAdaptiveGreedy2}, threshold $\frac{1}{2 \sqrt{k}}$ for \textsc{ContextualAdaptiveGreedy}. The turning point for \textsc{Explore-Then-Exploit} was set at $2,000$ for BibTeX, $4,000$ for Del.icio.us, and $12,000$ for Mediamill.

\begin{figure}[H]
\centering
\includegraphics[scale=0.55]{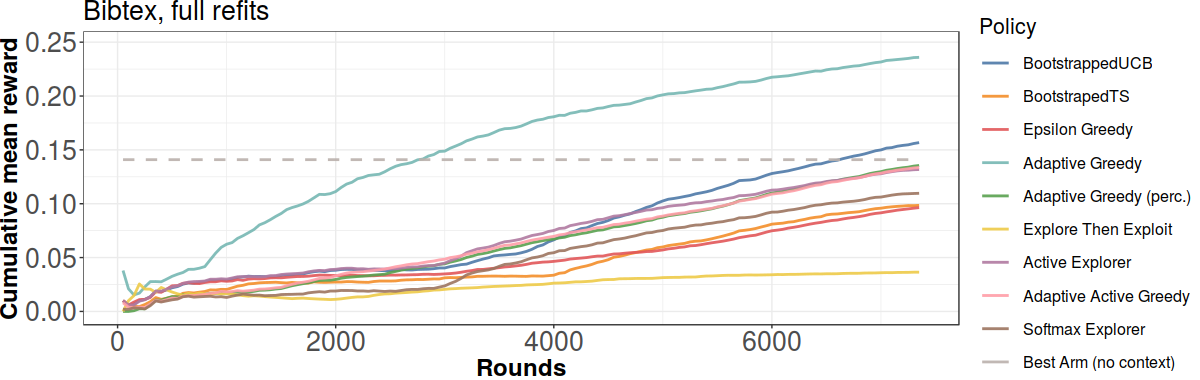}
\includegraphics[scale=0.55]{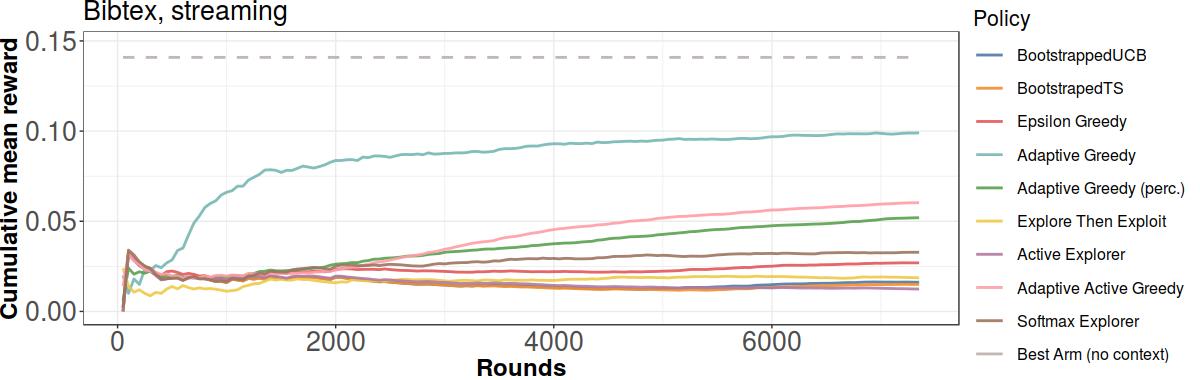}
\captionof{figure}{Results on BibTeX dataset}
\end{figure}

\begin{figure}[H]
\centering
\includegraphics[scale=0.53]{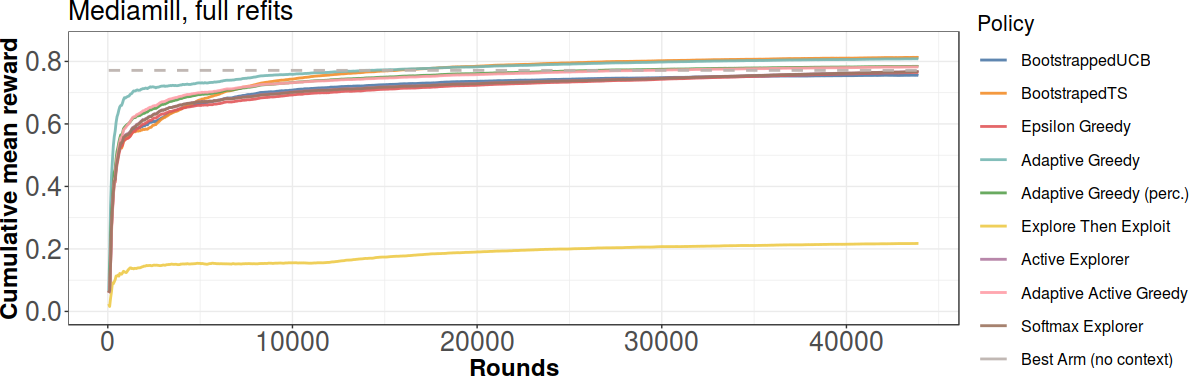}
\includegraphics[scale=0.53]{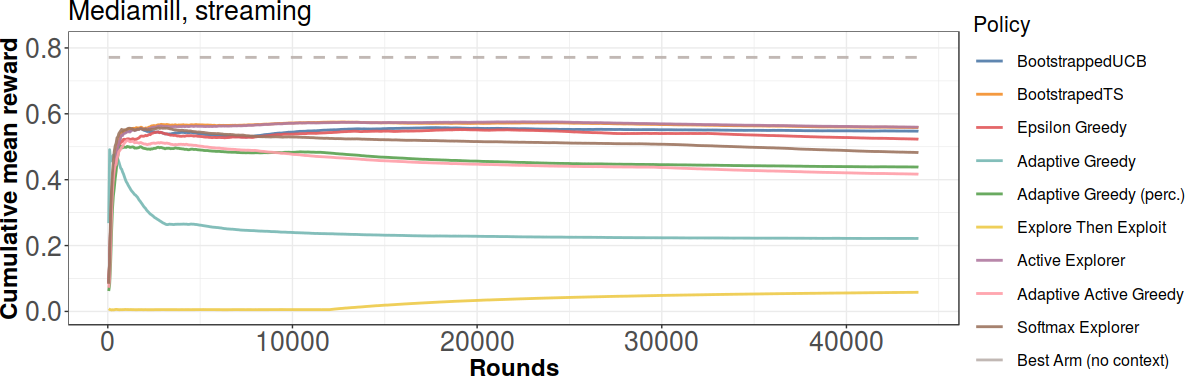}
\captionof{figure}{Results on Mediamill dataset}
\end{figure}

\begin{figure}[H]
\centering
\includegraphics[scale=0.53]{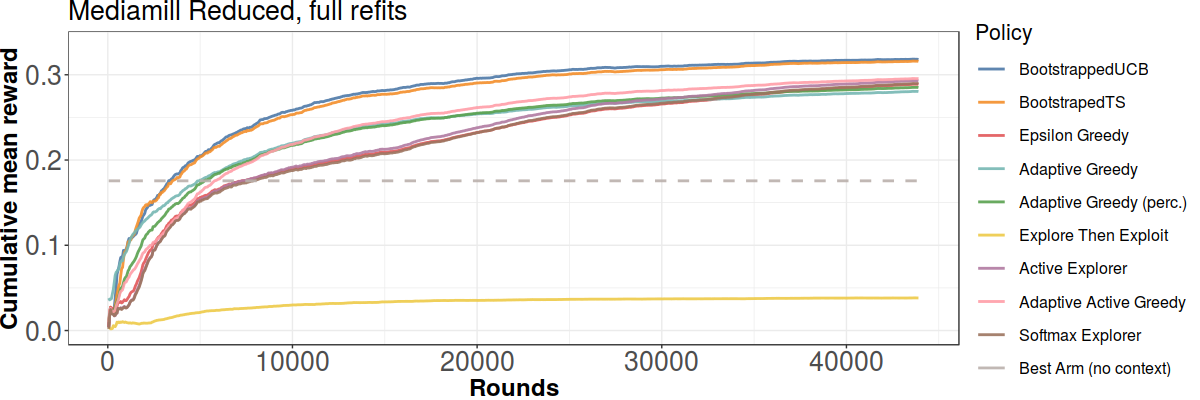}
\includegraphics[scale=0.53]{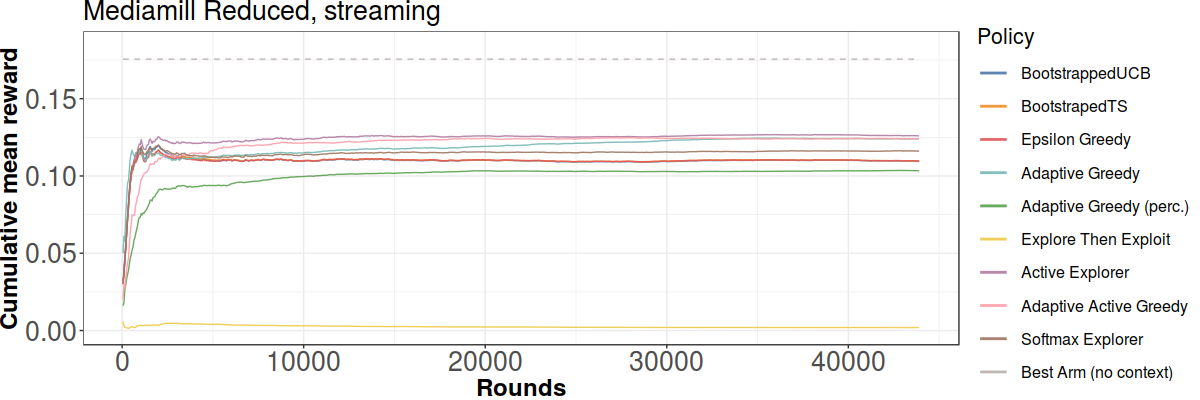}
\captionof{figure}{Results on Mediamill Reduced dataset}
\end{figure}

\begin{figure}[H]
\centering
\includegraphics[scale=0.55]{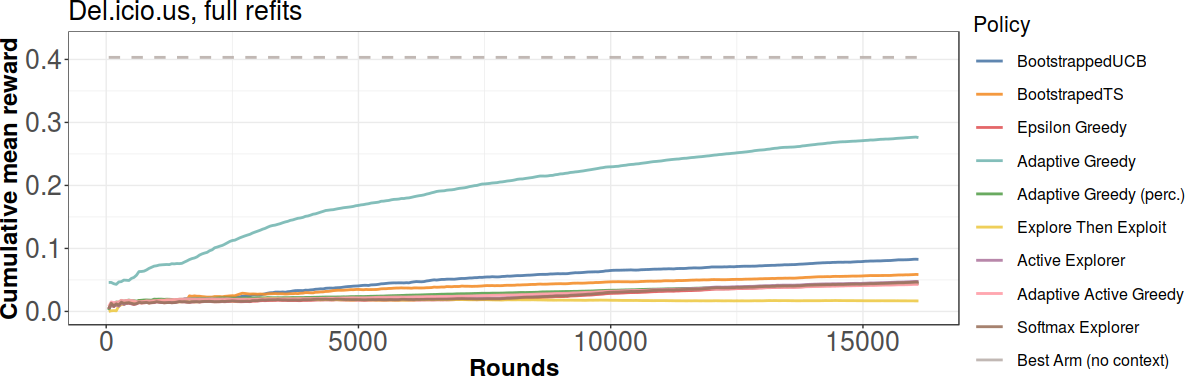}
\includegraphics[scale=0.55]{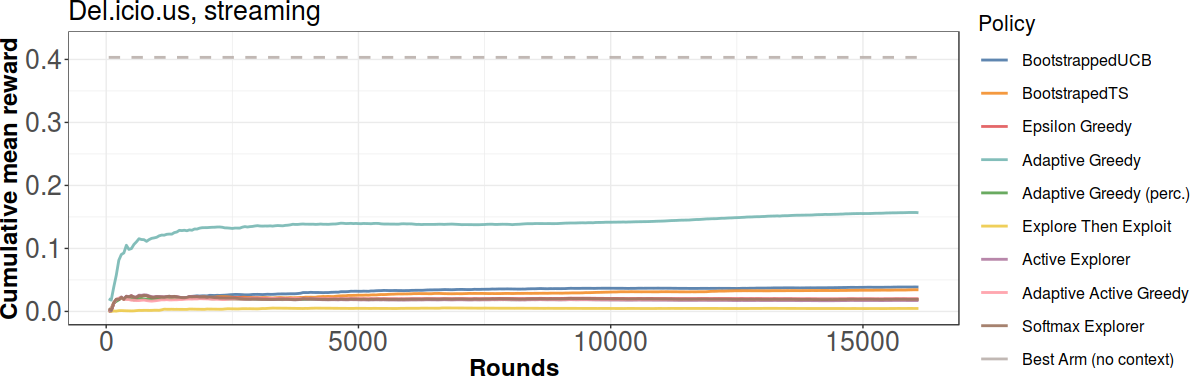}
\captionof{figure}{Results on Del.icio.us dataset}
\end{figure}

\section{Conclusions}

This work presented adaptations of the most common strategies or policies from multi-armed bandits to online contextual bandits scenarios with binary rewards through classification oracles.

Techniques such as bootstrapping or approximate bootstrapping were proposed for obtaining upper confidence bounds and for Thompson sampling, which resulted in more scalable approaches than previous works such as \textsc{LinUCB} (\cite{linucb1}), along with techniques that allow to start from zero rather than requiring an undefined earlier exploration phase like \cite{offsettree} or \cite{onlinecover}.

An empirical evaluation of adapted multi-armed bandits policies was performed, which in many cases showed better results compared to simpler baselines or to discarding the context. A further comparison with similar works meant for the regression setting was not feasible due to the lack of scalability of other algorithms.

Just like in MAB, the upper confidence bound approach proved to be a reasonably good strategy throughout all datasets despite the small number of resampes used, having fewer hyperparameters to tune. The overall best-performing policy however seems to be \textsc{ContextualAdaptiveGreedy}, which is also a faster approach. Enhancing it by incorporating active learning heuristics did not seem to have much of an effect, and it seems that seting a given initial threshold provides better results compared to setting the threshold as a moving percentile of the predictions.

While theoretically sound, using stochastic optimization to update classifiers with small batches of data resulted in severely degraded performance compared to full refits across all metaheuristics, even in the later rounds of larger datasets, with no policy managing to outperform choosing the best arm without context, at least with the hyperparameters experimented with for the \textsc{MAB-first} trick. This might in practice not be much a problem consider the short time it takes to fit models to a small number of observations as done in the earlier phases of a policy.

It shall be noted that all arms were treated as being independent from each other, which in reality might not be the case and other models incorporating similarity information might result in improved performance.

\bibliographystyle{plain}
\bibliography{cb}

\newpage
\begin{appendices}

\section{Online bootstrap for UCB}

In order to evaluate  how often the expected value of the target variable falls below the upper confidence bounds estimated by each method, simulations were performed at different sample sizes with randomly-generated data as follows:
\begin{itemize}
\item[$\bullet$ 1)] A set of coefficients is set as fixed (real coefficients), for all sample sizes and iterations.
\item[$\bullet$ 2)] Many iterations are performed in which covariates of a fixed sample size are generated at random, always from the same data-generating distribution, then the expected value of the target variable is calculated from the real coefficients, and random noise is added to it. This represents real samples from the data-generating process.
\item[$\bullet$ 3)] A test set is generated the same way as the samples in 2), but without adding noise to the target variable. This represents the real expected values.
\item[$\bullet$ 4)] Upper confidence bounds are generated for the observations in the test set under each real sample through the resampling methods described before.
\item[$\bullet$ 5)] The number of times that the real expected values are lower than the estimated upper bounds is calculated over all samples by taking the proportion of cases in the test set that were lower than the estimated bound.
\end{itemize}

Classification problems were also simulated similarly, but sampling the value for $y$ from a Bernoulli distribution instead, after applying a logistic transformation.

The confidence bound was defined at 80\% with number of resamples set at 10, which is a rather low and insufficient number, but is the kind of number that could be used without much of a speed penalty for the algorithms described in this work. In general, the statistic of interest in bootstrapping tends to be the standard error of coefficients, but in this case the estimations of expected values of the target variable are more directly related to the problem.

Another perhaps intuitive choice of random weighting would be $\text{Uniform}(0,1)$, but as can be seen, it severly underestimates the bounds. One might also think that less-variable Gamma weights could also work better at small sample sizes at the expense of worse results in larger sizes, so weights $\text{Gamma}(2,2)$ were also evaluated, but turned out to provide rather similar instability as the others in the proportion of estimations falling below the confidence bound. The $\text{Gamma}(1,1)$ weights provide almost the same results on average as the real bootstrap and the $\text{Poisson}(1)$ number of samples, and is perhaps a better alternative for classification as it avoids bad resamples resulting in only observations of one class. For a more intuitive explanation of the choice, recall that a $\text{Gamma}(1,1)$ distribution would indicate, for example, expected time between events that happen at rate $\sim \text{Poisson}(1)$, thereby acting as a mostly equivalent but smoother weighting than full inclusion/exclusion.

\begin{table}[H]
\caption {Proportion of expected values falling below 80\% confidence bound, Linear regression with large bias term, independent features}
\begin{adjustbox}{max width=\textwidth}{\centering
\begin{tabular}{|r|c|c|c|c|c|c|c|c|c|c|}
 \hline
 \textbf{Sample size} & \multicolumn{2}{|c|}{\textbf{Bootstrap}} & \multicolumn{2}{|c|}{\textbf{Poisson(1)}} & \multicolumn{2}{|c|}{\textbf{Uniform(0,1)}} & \multicolumn{2}{|c|}{\textbf{Gamma(1,1)}} & \multicolumn{2}{|c|}{\textbf{Gamma(2,2)}} \\
 & Mean & Std & Mean & Std & Mean & Std & Mean & Std & Mean & Std \\
 \hline
10 & 77.75\% & 20.87\% & 73.41\% & 19.54\% & 65.24\% & 21.75\% & 70.48\% & 20.72\% & 61.68\% & 22.36\% \\ 
15 & 77.67\% & 20.79\% & 75.20\% & 19.17\% & 62.60\% & 24.69\% & 67.79\% & 24.19\% & 62.57\% & 23.09\% \\ 
25 & 76.54\% & 20.04\% & 73.07\% & 21.12\% & 60.12\% & 22.13\% & 67.08\% & 20.73\% & 68.12\% & 21.09\% \\ 
39 & 74.37\% & 21.90\% & 73.53\% & 18.71\% & 67.92\% & 21.20\% & 73.70\% & 19.86\% & 66.32\% & 23.69\% \\ 
63 & 74.16\% & 21.96\% & 73.68\% & 21.55\% & 66.72\% & 24.24\% & 74.71\% & 21.63\% & 67.92\% & 23.24\% \\ 
100 & 75.39\% & 18.85\% & 75.60\% & 19.68\% & 67.29\% & 20.85\% & 75.64\% & 19.56\% & 68.02\% & 22.11\% \\ 
158 & 73.78\% & 19.88\% & 70.47\% & 22.39\% & 67.00\% & 21.41\% & 76.61\% & 19.78\% & 65.79\% & 22.76\% \\ 
251 & 75.12\% & 18.28\% & 74.77\% & 19.46\% & 66.45\% & 21.24\% & 75.47\% & 19.09\% & 67.51\% & 21.86\% \\ 
398 & 79.24\% & 19.79\% & 71.93\% & 21.45\% & 67.11\% & 19.30\% & 76.09\% & 17.79\% & 68.88\% & 21.00\% \\ 
630 & 79.29\% & 18.66\% & 76.65\% & 19.55\% & 66.99\% & 21.74\% & 75.29\% & 19.62\% & 68.16\% & 20.39\% \\ 
1,000 & 69.83\% & 23.56\% & 71.45\% & 20.54\% & 64.72\% & 21.88\% & 74.10\% & 19.01\% & 70.79\% & 22.67\% \\ 
1,584 & 74.05\% & 19.33\% & 74.66\% & 22.29\% & 64.10\% & 21.55\% & 74.00\% & 18.17\% & 68.22\% & 24.89\% \\ 
2,511 & 78.08\% & 19.21\% & 75.64\% & 18.86\% & 66.53\% & 21.97\% & 75.67\% & 19.96\% & 63.53\% & 22.12\% \\ 
3,981 & 74.39\% & 20.25\% & 75.77\% & 22.12\% & 65.25\% & 26.60\% & 73.22\% & 22.92\% & 69.46\% & 21.84\% \\ 
6,309 & 75.04\% & 21.68\% & 74.48\% & 17.74\% & 63.86\% & 23.60\% & 74.05\% & 21.45\% & 70.01\% & 23.09\% \\ 
10,000 & 77.04\% & 21.72\% & 76.68\% & 18.99\% & 67.13\% & 22.47\% & 76.39\% & 19.69\% & 70.14\% & 20.58\% \\ \hline
 \hline
\end{tabular}}\end{adjustbox}
\begin{tablenotes}
      \small
      \item  The model used was $y = 8 + 1.05 x_1 - 2.35 x_2 + 0.15 x_3 + \epsilon$, with $x_1, x_2, x_3, \epsilon \sim N(0,1)$, 100 samples, 10 resamples per sample, test sample size $n=1000$.
    \end{tablenotes}
\end{table}

\begin{table}[H]
\caption {Proportion of expected values falling below 80\% confidence bound, Logistic regression with small negative bias term, independent features}
\begin{adjustbox}{max width=\textwidth}{\centering
\begin{tabular}{|r|c|c|c|c|c|c|c|c|c|c|}
 \hline
 \textbf{Sample size} & \multicolumn{2}{|c|}{\textbf{Bootstrap}} & \multicolumn{2}{|c|}{\textbf{Poisson(1)}} & \multicolumn{2}{|c|}{\textbf{Uniform(0,1)}} & \multicolumn{2}{|c|}{\textbf{Gamma(1,1)}} & \multicolumn{2}{|c|}{\textbf{Gamma(2,2)}} \\
 & Mean & Std & Mean & Std & Mean & Std & Mean & Std & Mean & Std \\
 \hline
10 & 85.36\% & 6.60\% & 84.10\% & 6.81\% & 79.64\% & 6.25\% & 80.93\% & 6.42\% & 80.96\% & 6.13\% \\
15 & 81.36\% & 5.08\% & 83.84\% & 4.39\% & 80.49\% & 4.48\% & 81.57\% & 4.63\% & 80.30\% & 4.33\% \\
25 & 81.73\% & 3.59\% & 81.08\% & 4.30\% & 79.44\% & 3.65\% & 80.32\% & 3.82\% & 79.95\% & 3.19\% \\
39 & 80.27\% & 2.48\% & 80.27\% & 2.91\% & 79.64\% & 2.64\% & 80.43\% & 2.75\% & 79.14\% & 2.69\% \\
63 & 79.78\% & 2.01\% & 79.61\% & 2.33\% & 79.09\% & 2.09\% & 79.74\% & 2.11\% & 79.55\% & 1.90\% \\
100 & 79.27\% & 1.58\% & 79.51\% & 1.70\% & 78.68\% & 1.62\% & 79.33\% & 1.63\% & 79.05\% & 1.64\% \\
158 & 79.21\% & 1.09\% & 79.05\% & 1.29\% & 78.80\% & 1.13\% & 79.30\% & 1.20\% & 78.54\% & 1.19\% \\
251 & 78.89\% & 1.04\% & 79.10\% & 1.05\% & 78.54\% & 0.96\% & 78.95\% & 0.99\% & 78.67\% & 0.95\% \\
398 & 78.77\% & 0.72\% & 78.79\% & 0.92\% & 78.56\% & 0.99\% & 78.87\% & 0.99\% & 78.44\% & 0.87\% \\
630 & 78.50\% & 0.80\% & 78.67\% & 0.68\% & 78.34\% & 0.73\% & 78.59\% & 0.74\% & 78.48\% & 0.57\% \\
1000 & 78.52\% & 0.64\% & 78.45\% & 0.59\% & 78.22\% & 0.50\% & 78.45\% & 0.53\% & 78.33\% & 0.56\% \\
1584 & 78.45\% & 0.43\% & 78.39\% & 0.46\% & 78.30\% & 0.43\% & 78.48\% & 0.43\% & 78.31\% & 0.43\% \\
2511 & 78.31\% & 0.36\% & 78.31\% & 0.37\% & 78.23\% & 0.37\% & 78.38\% & 0.35\% & 78.20\% & 0.39\% \\
3981 & 78.37\% & 0.29\% & 78.34\% & 0.29\% & 78.21\% & 0.33\% & 78.32\% & 0.34\% & 78.26\% & 0.29\% \\
6309 & 78.22\% & 0.26\% & 78.25\% & 0.25\% & 78.18\% & 0.23\% & 78.27\% & 0.23\% & 78.24\% & 0.23\% \\
10000 & 78.24\% & 0.20\% & 78.26\% & 0.19\% & 78.19\% & 0.19\% & 78.28\% & 0.20\% & 78.23\% & 0.17\% \\ \hline
 \hline
\end{tabular}}\end{adjustbox}
\begin{tablenotes}
      \small
      \item  The model used was $y \sim \text{Bernoulli}(\frac{1}{1 + \exp(-2 + 1.05 x_1 - 2.35 x_2 + 0.15 x_3 + \epsilon)}$, with $x_1, x_2, x_3, \sim N(0, 0.5)$, $\epsilon  \sim N(0, 0.5)$, 100 samples, 10 resamples per sample, test sample size $n=1000$.
    \end{tablenotes}
\end{table}

\begin{table}[H]
\caption {Proportion of expected values falling below 80\% confidence bound, Logistic regression with small negative bias term, correlated features}
\begin{adjustbox}{max width=\textwidth}{\centering
\begin{tabular}{|r|c|c|c|c|c|c|c|c|c|c|}
 \hline
 \textbf{Sample size} & \multicolumn{2}{|c|}{\textbf{Bootstrap}} & \multicolumn{2}{|c|}{\textbf{Poisson(1)}} & \multicolumn{2}{|c|}{\textbf{Uniform(0,1)}} & \multicolumn{2}{|c|}{\textbf{Gamma(1,1)}} & \multicolumn{2}{|c|}{\textbf{Gamma(2,2)}} \\
 & Mean & Std & Mean & Std & Mean & Std & Mean & Std & Mean & Std \\
 \hline
10 & 71.44\% & 3.70\% & 73.09\% & 5.27\% & 70.20\% & 3.04\% & 70.85\% & 3.16\% & 69.83\% & 3.43\% \\
15 & 70.70\% & 2.62\% & 70.90\% & 2.56\% & 69.89\% & 2.19\% & 70.52\% & 2.33\% & 69.80\% & 2.02\% \\
25 & 69.92\% & 1.74\% & 69.93\% & 1.66\% & 69.40\% & 1.55\% & 69.89\% & 1.66\% & 69.24\% & 1.43\% \\
39 & 69.61\% & 1.19\% & 69.71\% & 1.06\% & 69.22\% & 1.17\% & 69.56\% & 1.19\% & 69.15\% & 1.15\% \\
63 & 69.47\% & 0.82\% & 69.50\% & 0.90\% & 68.95\% & 0.82\% & 69.28\% & 0.82\% & 69.21\% & 0.84\% \\
100 & 69.18\% & 0.75\% & 69.38\% & 0.67\% & 68.92\% & 0.65\% & 69.21\% & 0.67\% & 69.07\% & 0.64\% \\
158 & 69.05\% & 0.54\% & 69.07\% & 0.55\% & 68.83\% & 0.47\% & 69.01\% & 0.50\% & 68.89\% & 0.45\% \\
251 & 69.08\% & 0.40\% & 68.97\% & 0.38\% & 68.80\% & 0.30\% & 68.94\% & 0.33\% & 68.89\% & 0.38\% \\
398 & 68.92\% & 0.31\% & 68.96\% & 0.32\% & 68.80\% & 0.26\% & 68.91\% & 0.28\% & 68.82\% & 0.25\% \\
630 & 68.89\% & 0.23\% & 68.85\% & 0.17\% & 68.78\% & 0.20\% & 68.87\% & 0.21\% & 68.79\% & 0.16\% \\
1000 & 68.81\% & 0.19\% & 68.80\% & 0.14\% & 68.77\% & 0.13\% & 68.82\% & 0.15\% & 68.78\% & 0.12\% \\
1584 & 68.80\% & 0.10\% & 68.78\% & 0.14\% & 68.79\% & 0.11\% & 68.83\% & 0.11\% & 68.76\% & 0.13\% \\
2511 & 68.80\% & 0.10\% & 68.81\% & 0.10\% & 68.77\% & 0.10\% & 68.81\% & 0.09\% & 68.79\% & 0.11\% \\
3981 & 68.82\% & 0.09\% & 68.81\% & 0.09\% & 68.79\% & 0.10\% & 68.82\% & 0.09\% & 68.79\% & 0.10\% \\
6309 & 68.82\% & 0.09\% & 68.81\% & 0.09\% & 68.78\% & 0.09\% & 68.81\% & 0.09\% & 68.80\% & 0.09\% \\
10000 & 68.82\% & 0.08\% & 68.81\% & 0.09\% & 68.78\% & 0.09\% & 68.82\% & 0.08\% & 68.80\% & 0.09\% \\ \hline
 \hline
\end{tabular}}\end{adjustbox}
\begin{tablenotes}
      \small
      \item  The model used was $y \sim \text{Bernoulli}(\frac{1}{1 + \exp(-2 + 1.05 x_1 - 2.35 x_2 + 0.15 x_3 + \epsilon)}$, with $x_1, x_2, x_3, \sim N(\begin{bmatrix} 0, 0, 0\end{bmatrix}, \begin{bmatrix} 3.17 & -1.08 & -2.19 \\ -1.08 & 2.23 & 1.10 \\ -2.19 & 1.10 & 1.63 \end{bmatrix})$, $\epsilon  \sim N(0, 0.5)$, 100 samples, 10 resamples per sample, test sample size $n=1000$.
    \end{tablenotes}
\end{table}

The results are rather different in the case of independent and correlated components for logistic regression (note that it purposefully added extra noise to the Bernoulli parameter), but for linear regression (not shown here) they are pretty much the same. Although the numbers here suggest sistematic underestimation of the bounds regardless of the method, changing the bias term has a large effect on the numbers, e.g. using small bias (not shown here) results in the upper bounds being severly overestimated in the case of linear regression, and changing the variance of the random noise also leads to large changes in the variance of the estimated bounds.

While this small simulation does not constitute any rigurous proof, it suggests that random weights can be a more stable alternative for online bootstrapping than Poisson-distributed number of occurrences, without any loss of precision in the estimated bound.

\section{Smoothing and MAB-first, many-armed bandits}
Both the smoothing criterion and the \textsc{MAB-first} trick can help with estimations in arms that have seen few or no examples of the positive class for that classifier. These were compared with different values of their hyperparameters on the Bibtex, Eurlex and Mediamill datasets, using as metaheuristics \textsc{Epsilon-Greedy}, \textsc{BootstrappedTS} and \textsc{BootstrappedUCB}. The base classifier used is logistic regression. These statistics represent just one run, rather than an average over multiple runs as in the other plots.

\begin{figure}[H]
\centering
\includegraphics[scale=0.55]{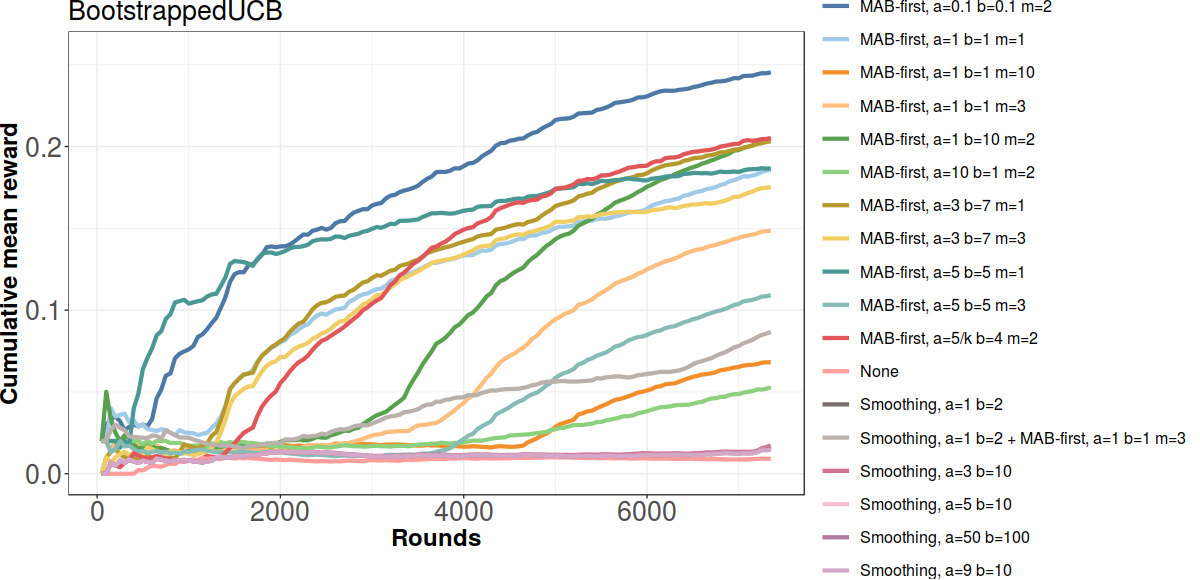}
\includegraphics[scale=0.55]{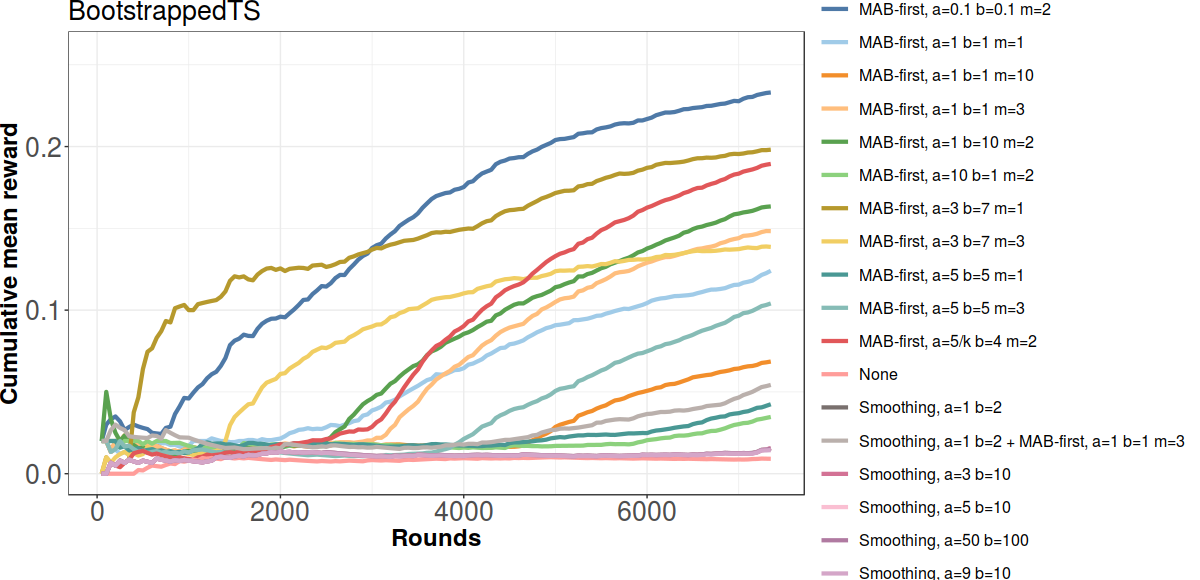}
\includegraphics[scale=0.55]{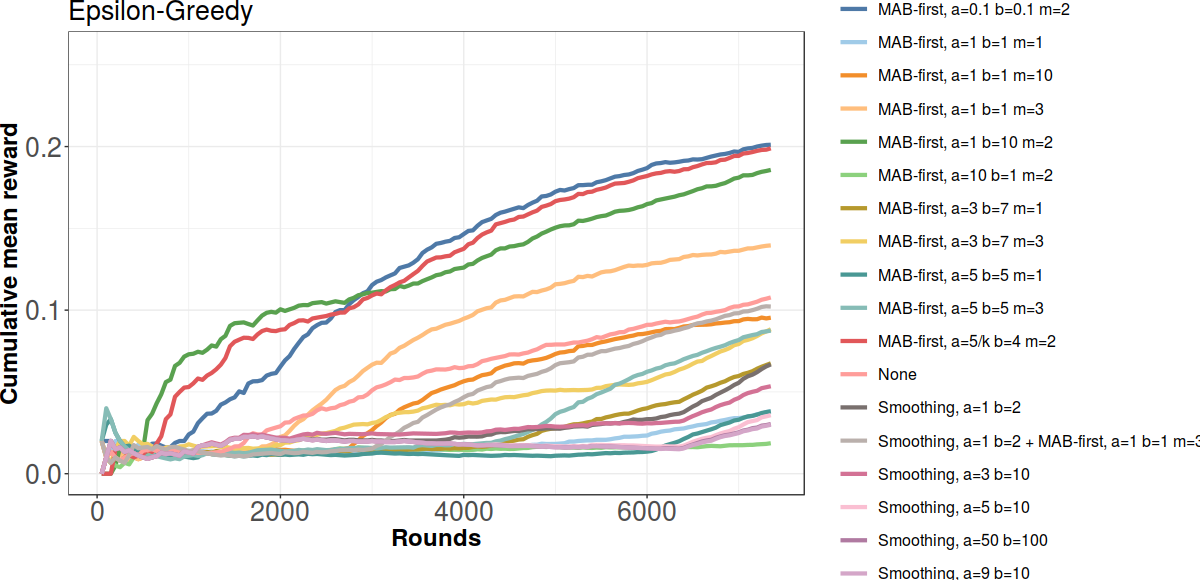}
\captionof{figure}{Smoothing and MAB-first on BibTeX dataset}
\end{figure}

\begin{figure}[H]
\centering
\includegraphics[scale=0.55]{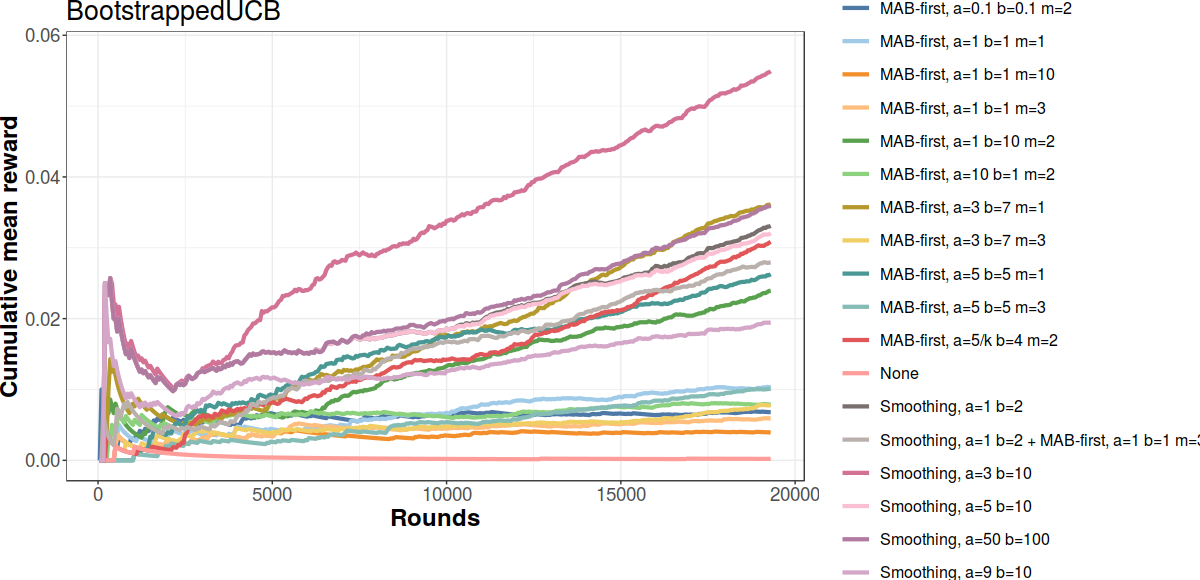}
\includegraphics[scale=0.55]{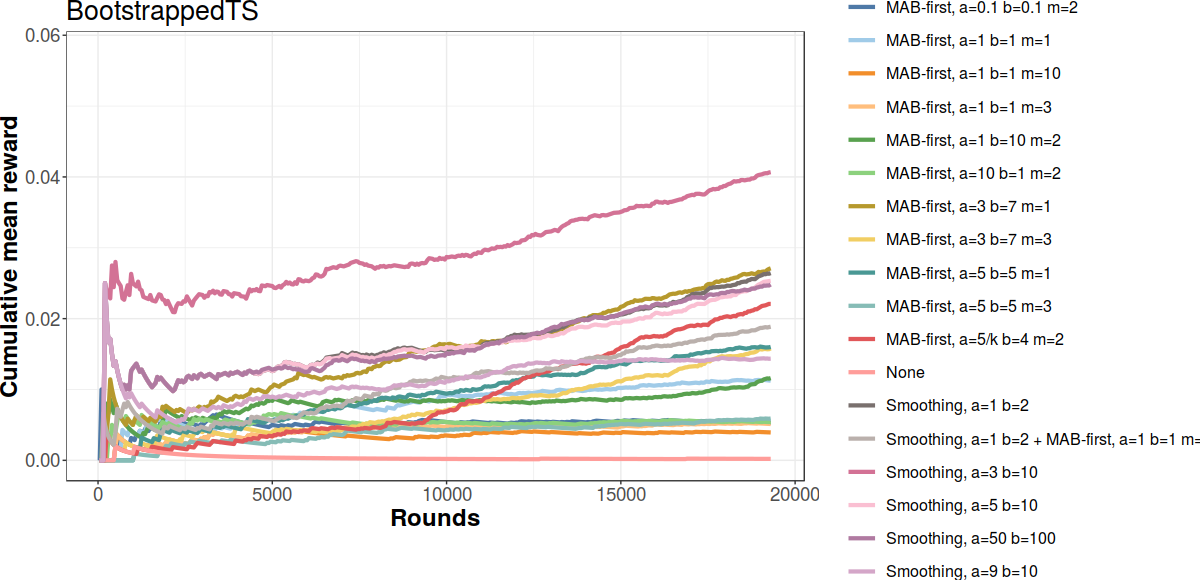}
\includegraphics[scale=0.55]{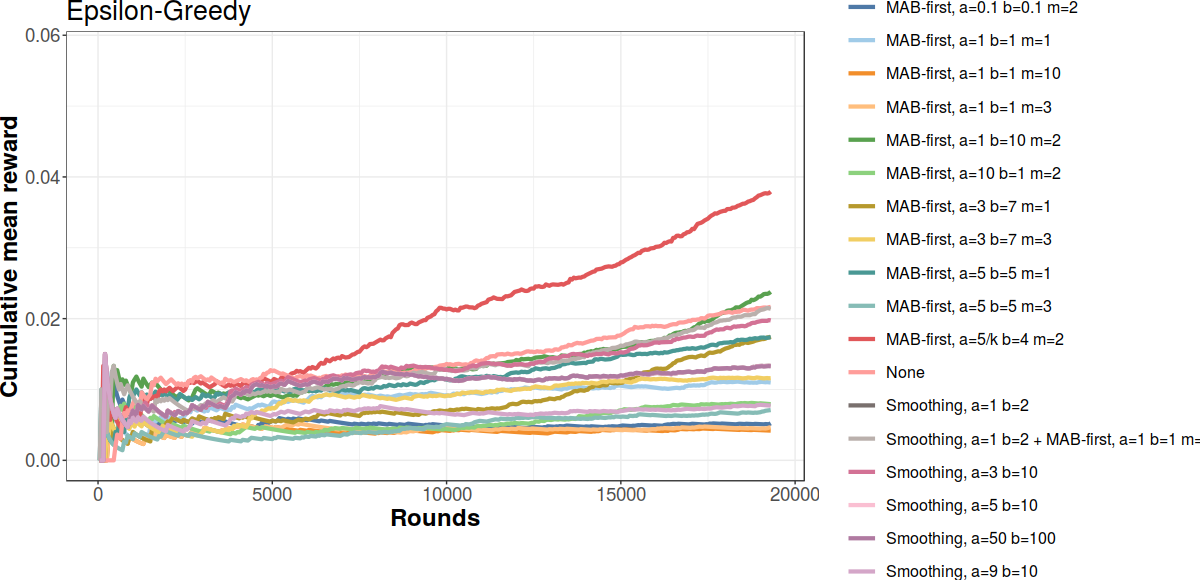}
\captionof{figure}{Smoothing and MAB-first on Eurlex dataset}
\end{figure}

\begin{figure}[H]
\centering
\includegraphics[scale=0.55]{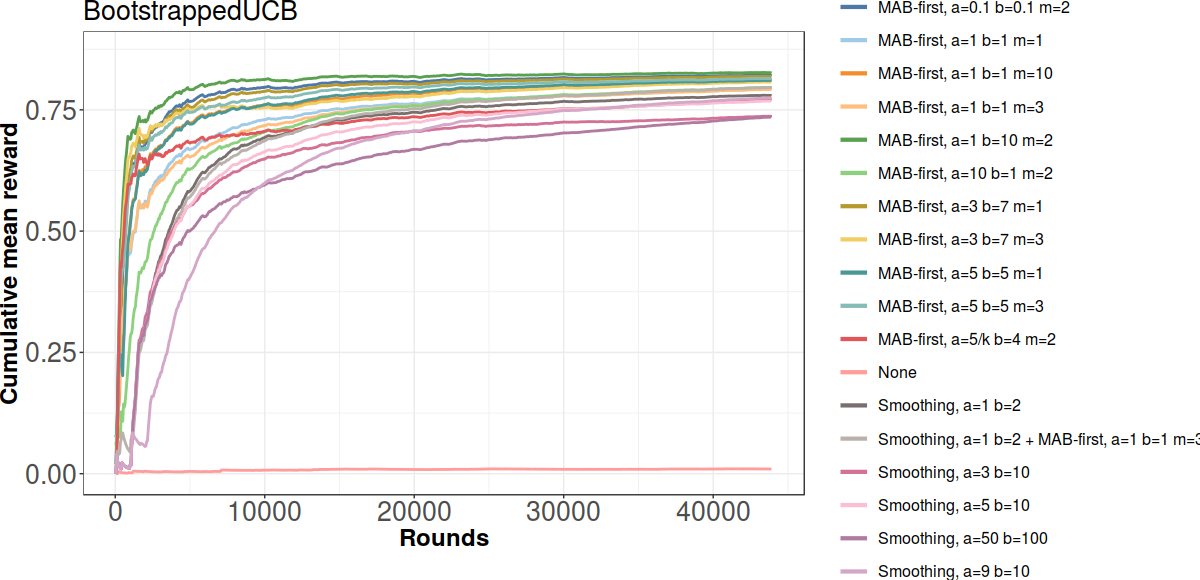}
\includegraphics[scale=0.55]{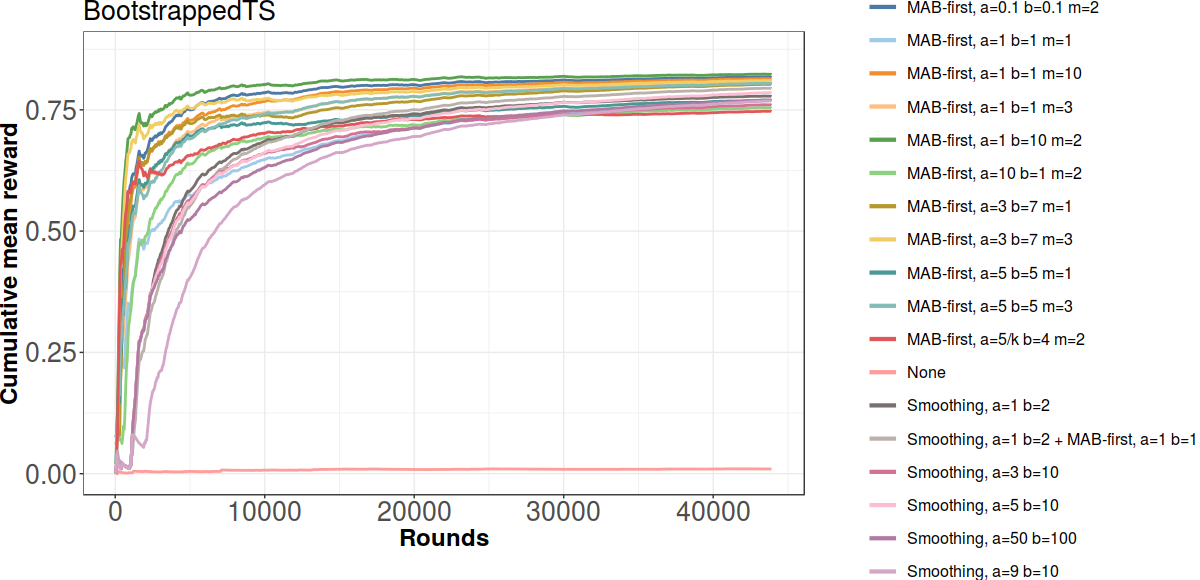}
\includegraphics[scale=0.55]{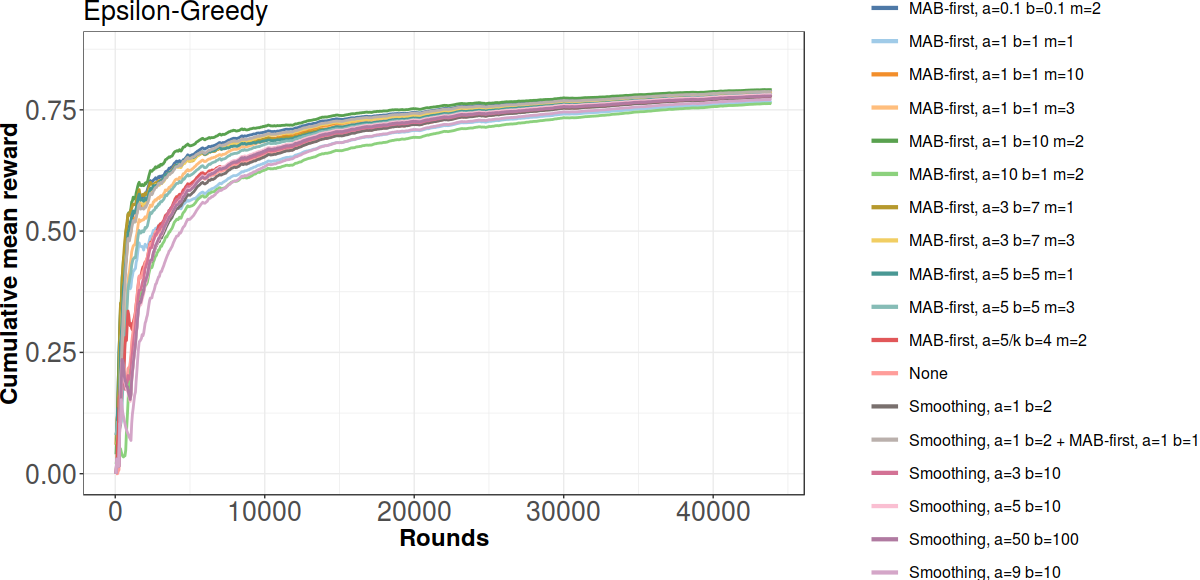}
\captionof{figure}{Smoothing and MAB-first on Mediamill dataset}
\end{figure}

\begin{figure}[H]
\centering
\includegraphics[scale=0.55]{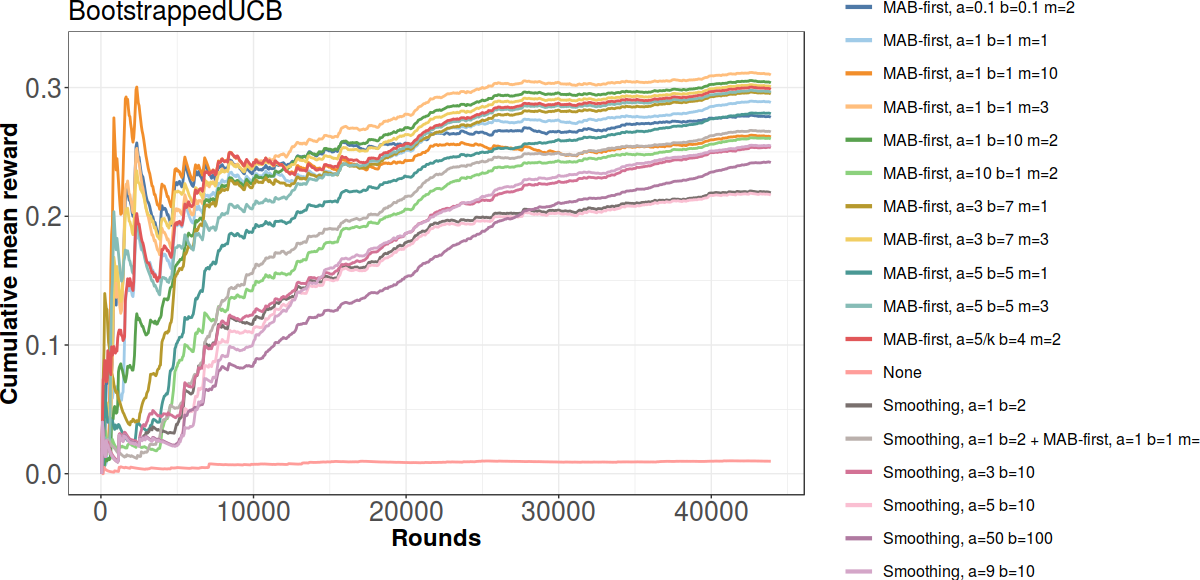}
\includegraphics[scale=0.55]{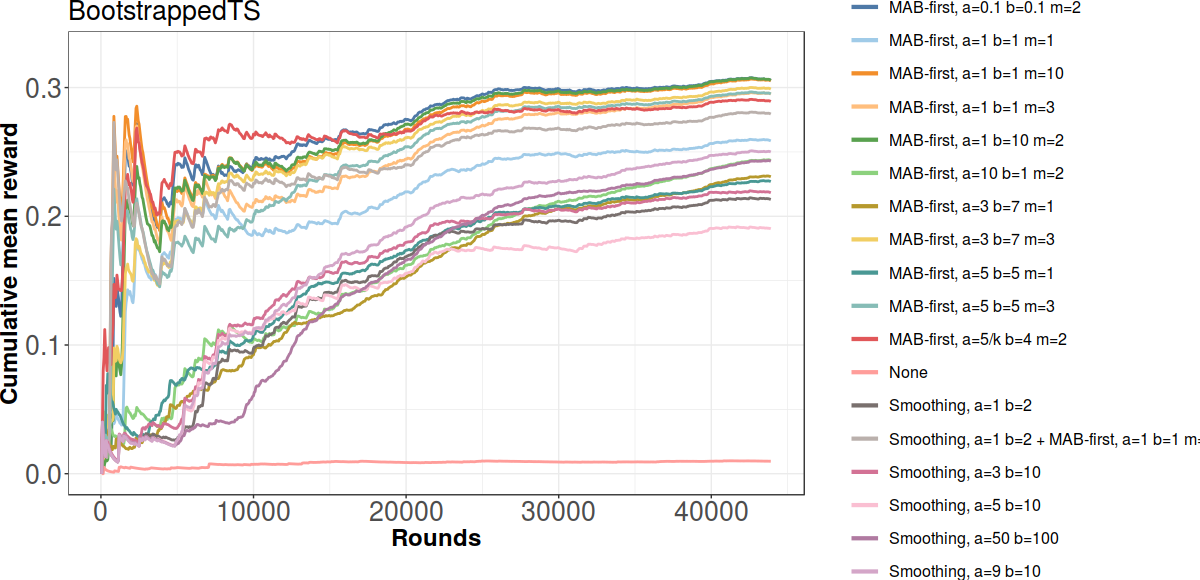}
\includegraphics[scale=0.55]{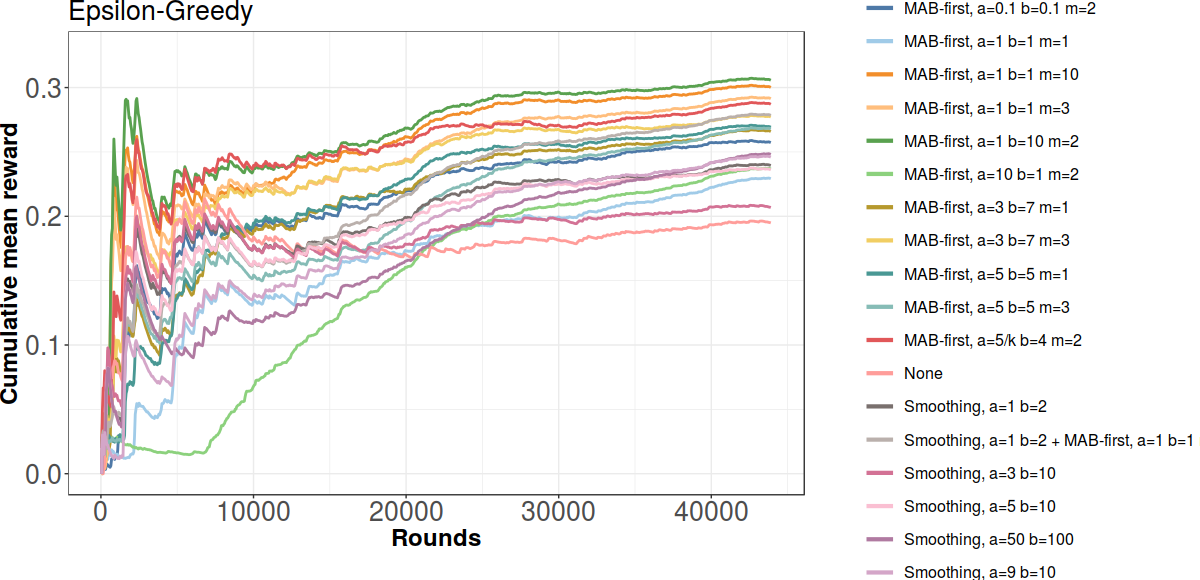}
\captionof{figure}{Smoothing and MAB-first on Mediamill Reduced dataset}
\end{figure}

The number of arms likely plays a role in the effectiveness of these heuristics. It's reasonable to think that limiting the number of arms would bring better results when the number of rounds is not infinite, so experiments were run limiting the number of arms to random subsets of varying cardinality. If there is a dominant arm that tends to perform better than the others and it doesn't get included in the subset being used, performance should suffer significantly - this is precisely the case in the Mediamill dataset.

\begin{figure}[H]
\centering
\includegraphics[scale=0.53]{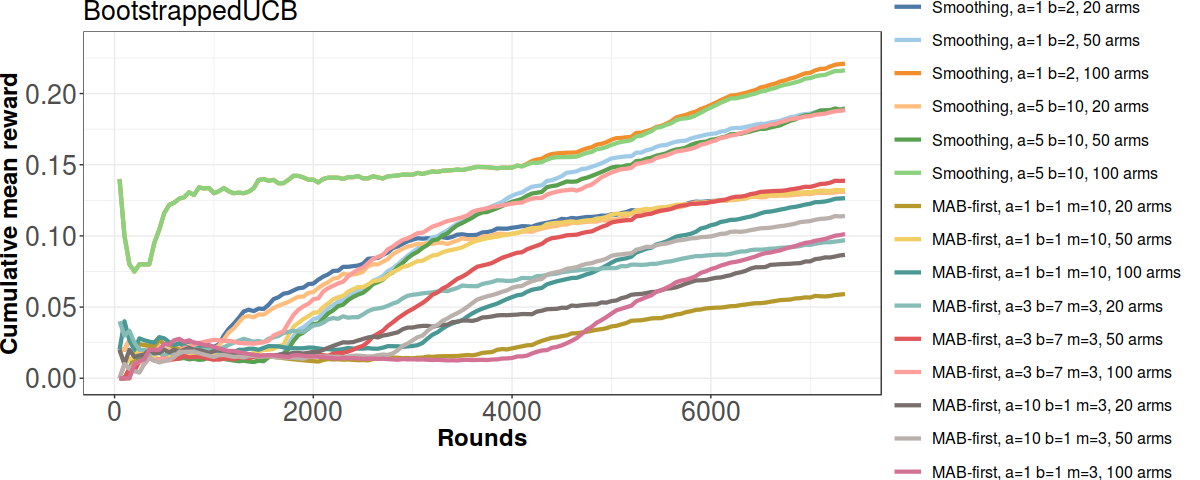}
\includegraphics[scale=0.53]{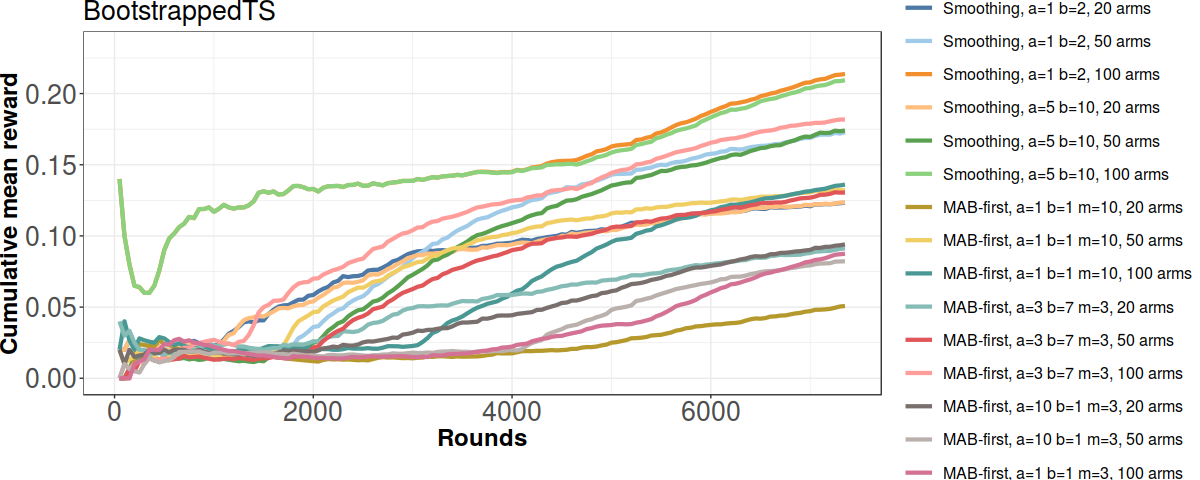}
\includegraphics[scale=0.53]{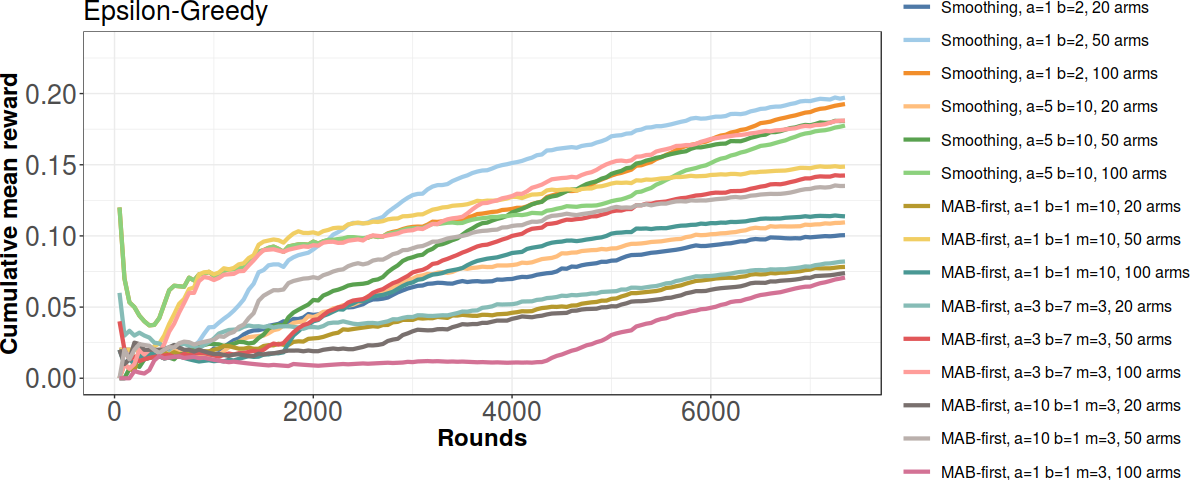}
\captionof{figure}{Limiting the number of arms, Bibtex dataset}
\end{figure}

\begin{figure}[H]
\centering
\includegraphics[scale=0.55]{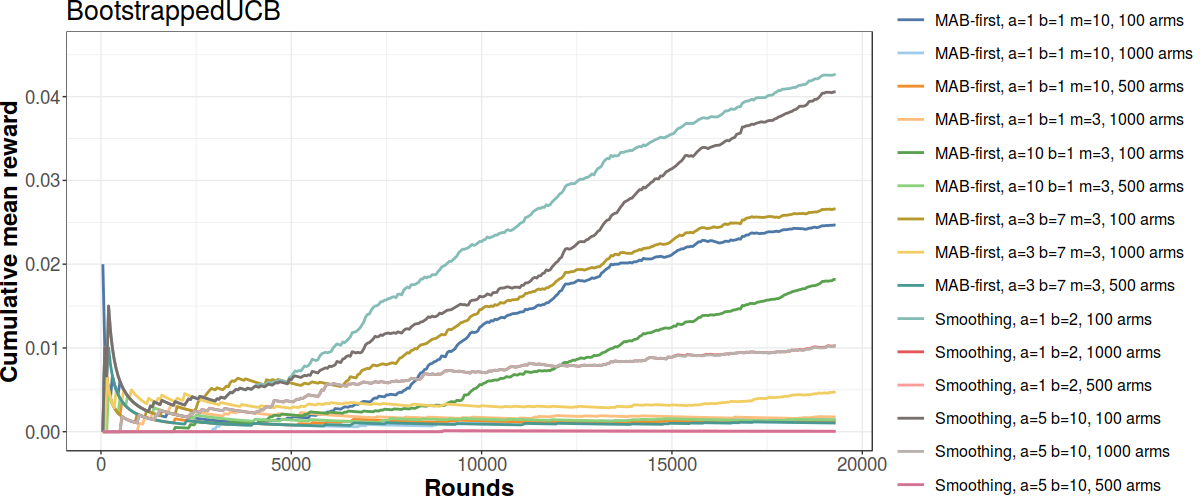}
\includegraphics[scale=0.55]{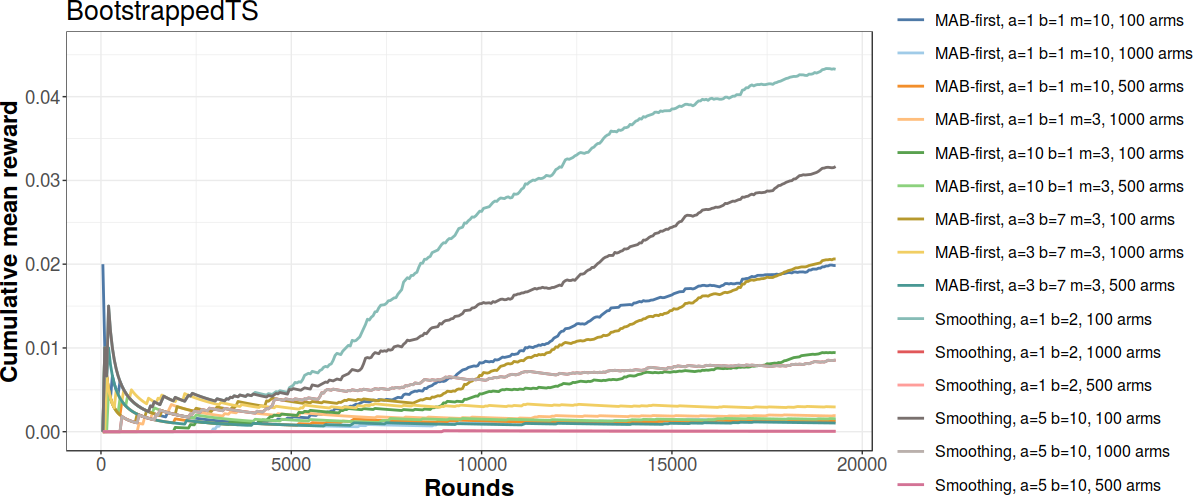}
\includegraphics[scale=0.55]{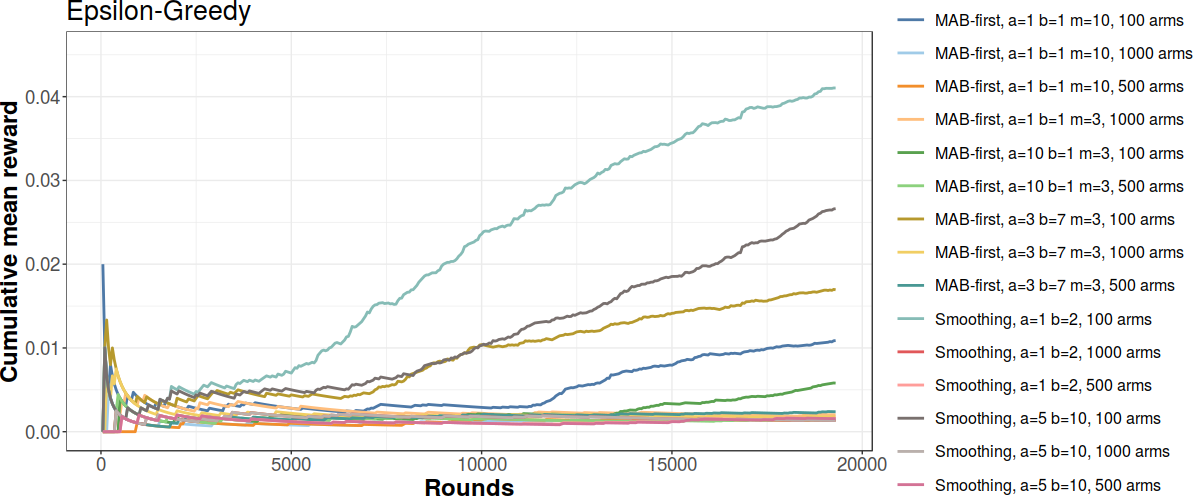}
\captionof{figure}{Limiting the number of arms, Eurlex dataset}
\end{figure}

\begin{figure}[H]
\centering
\includegraphics[scale=0.55]{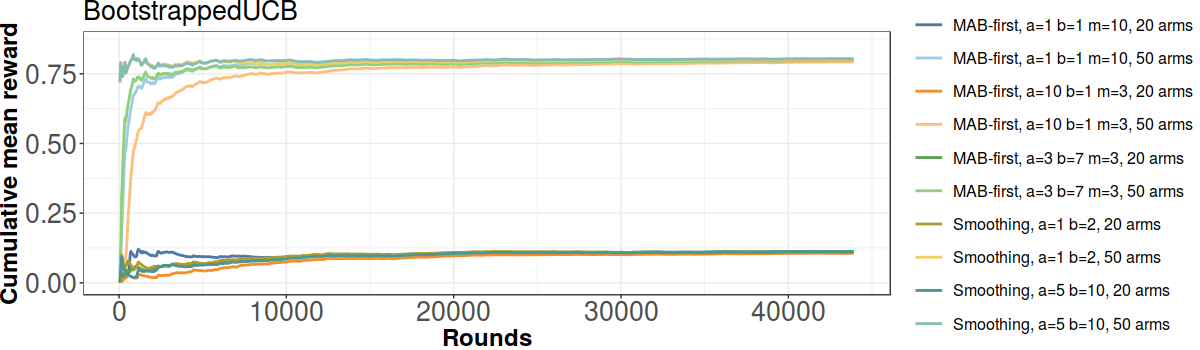}
\includegraphics[scale=0.55]{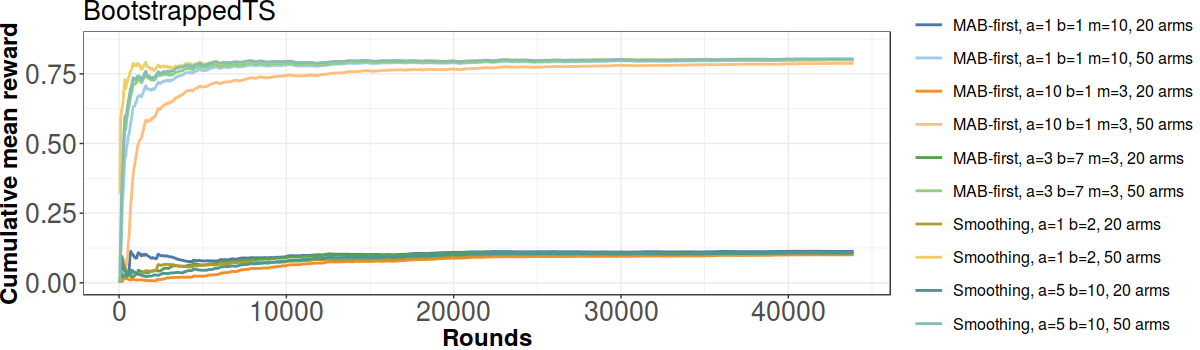}
\includegraphics[scale=0.55]{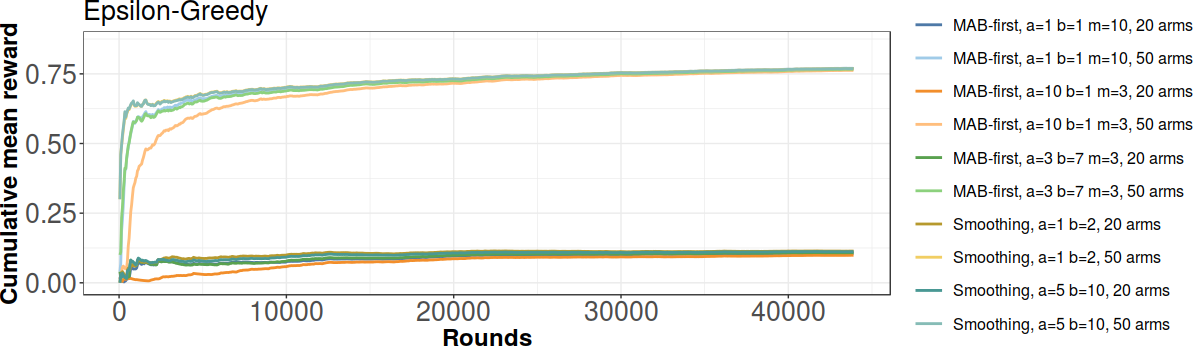}
\captionof{figure}{Limiting the number of arms, Mediamill dataset}
\end{figure}

\begin{figure}[H]
\centering
\includegraphics[scale=0.55]{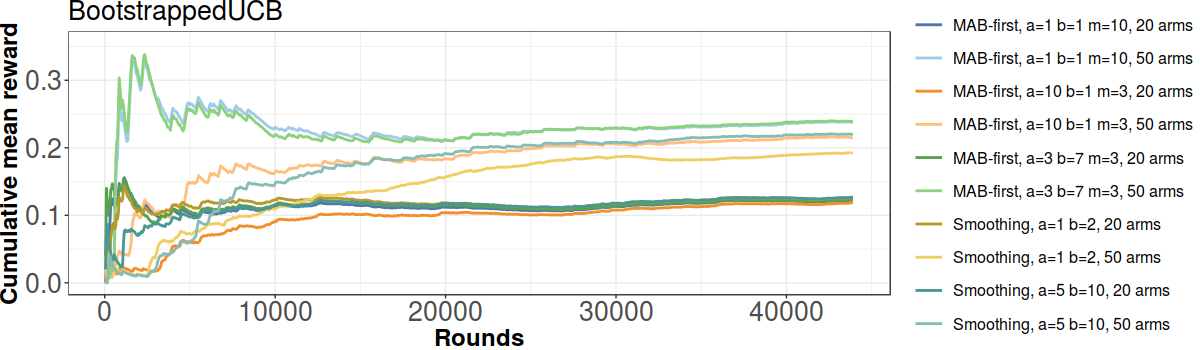}
\includegraphics[scale=0.55]{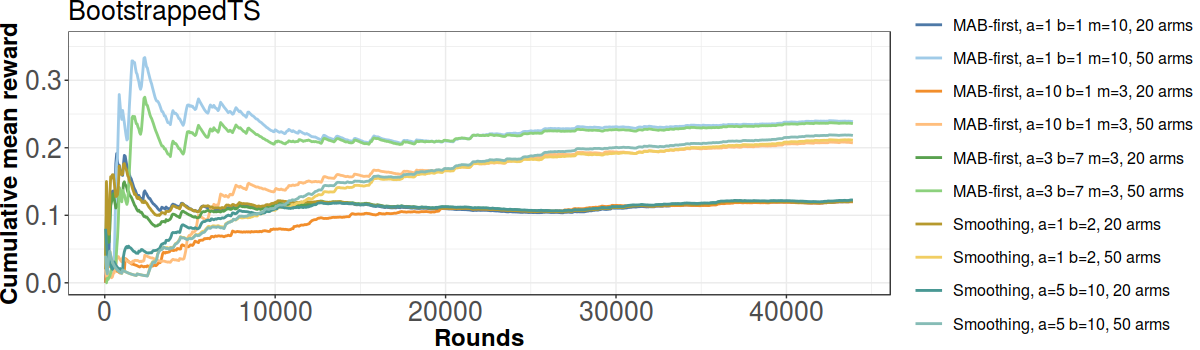}
\includegraphics[scale=0.55]{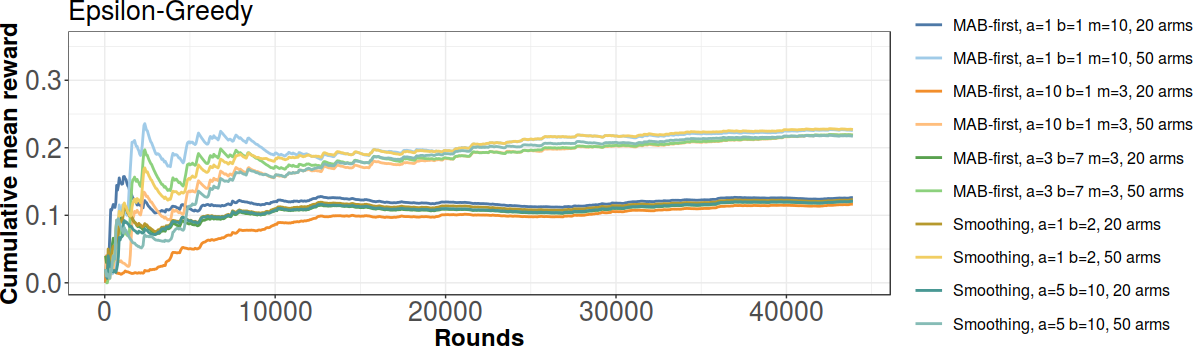}
\captionof{figure}{Limiting the number of arms, Mediamill Reduced dataset}
\end{figure}

It's not possible to conclude that one approach is always better than the other, but \textsc{MAB-first} seems to have an edge. It might be the case that smoothing works better in situations in which the number of arms is very large, but it's not possible to generalize from only these datasets.

The choice of hyperparameters for both the smoothing and \textsc{MAB-first} has a very large impact on the results, perhaps even more than the selection of contextual bandit strategy. Nevertheless, regardless of the choice of hyperparameters, they both provide a significant improvement compared to not using them at all when the contextual bandit policy does not choose arms uniformly at random.

These results seem to indicate that, for \textsc{BootstrappedUCB} and \textsc{BootstrappedTS}, when the choice of hyperparameters for \textsc{MAB-first} and smoothing is good, the more arms that are considered, the better the results, despite the relatively large number of arms compared to the number of rounds in these experiments. For \textsc{Epsilon-Greedy}, there seemed to be a lot more variance in this regard, but overall, a larger number of available arms also seems to lead to better results.

\section{Gradient norms in active learning}
A small comparison of the \textsc{ActiveExplorer} and the \textsc{ActiveAdaptiveGreedy} algorithms using as active learning selection heuristic either the minimum, maximum, or weighted average of the gradient norms under each label for a given observation was done on the Bibtex dataset. Just like in the main comparison, the base classifier used is logistic regression, with the \textsc{MAB-first} trick, and models refit to the full dataset every 50 rounds. The comparison is an average of 10 runs with the dataset shuffled differently in each.

\begin{figure}[H]%
    \centering
    \subfloat{{\includegraphics[width=7cm]{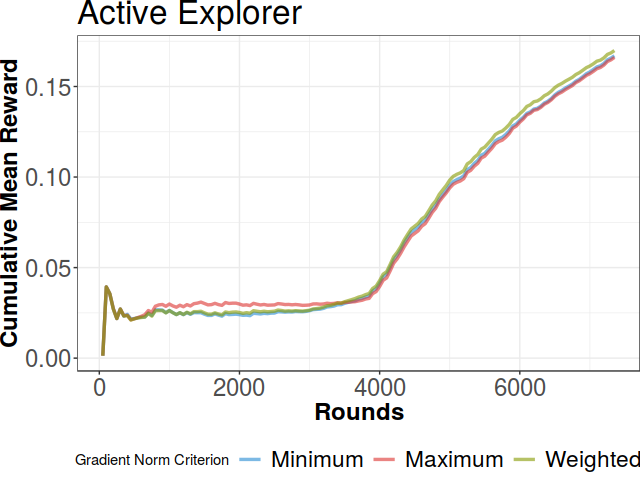} }}%
    \qquad
    \subfloat{{\includegraphics[width=7cm]{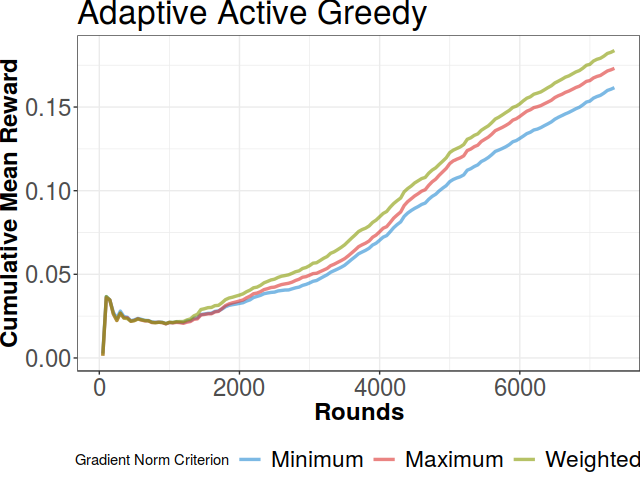} }}%
    \caption{Comparison of active learning selection criteria on Bibtex dataset}%
    \label{fig:example}%
\end{figure}

As can be seen, using the weighted average of the norms (as described in the algorithm) provides slightly better results under both metaheuristics experimented with compared to using the minimum or the maximum of either label.

\end{appendices}
\end{document}